\documentclass[10pt]{article}

\usepackage{microtype}
\usepackage{graphicx}
\usepackage{booktabs} 
\usepackage{pdfpages}
\usepackage{caption}
\usepackage{tabularx}
\usepackage{subcaption}
\usepackage{longtable}
\usepackage{pdflscape}
\usepackage{algorithm}
\usepackage{algorithmicx}
\usepackage{algpseudocode}
\usepackage{multirow}
\usepackage{diagbox}
\usepackage{subfloat}
\usepackage{verbatimbox}
\usepackage{float}
\usepackage{array}
\usepackage{threeparttable}

\usepackage{hyperref}
\usepackage{enumitem}

\usepackage[accepted]{icml2024}

\usepackage{amsmath}
\usepackage{amssymb}
\usepackage{mathtools}
\usepackage{amsthm}
\usepackage{color} 
\usepackage{amsfonts}
\usepackage{booktabs}
\usepackage{xtab}
\usepackage{afterpage}
\usepackage[capitalize,noabbrev]{cleveref}
\usepackage{diagbox}

\theoremstyle{plain}
\newtheorem{theorem}{Theorem}[section]
\newtheorem{proposition}[theorem]{Proposition}
\newtheorem{lemma}[theorem]{Lemma}

\theoremstyle{definition}

\theoremstyle{remark}

\usepackage[textsize=tiny]{todonotes}

\icmltitlerunning{Conformal Prediction for Deep Classifier via Label Ranking}

\begin{document}

\twocolumn[
\icmltitle{Conformal Prediction for Deep Classifier via Label Ranking}



\icmlsetsymbol{equal}{*}
\icmlsetsymbol{icmlWorkDone}{†}

\begin{icmlauthorlist}
\icmlauthor{Jianguo Huang}{equal,a,b,icmlWorkDone}
\icmlauthor{Huajun Xi}{equal,a}
\icmlauthor{Linjun Zhang}{c}
\icmlauthor{Huaxiu Yao}{d}
\icmlauthor{Yue Qiu}{e}
\icmlauthor{Hongxin Wei}{a}
\end{icmlauthorlist}

\icmlaffiliation{a}{Department of Statistics and Data Science, Southern University of Science and Technology.}
\icmlaffiliation{b}{The School of Information Science and Technology, ShanghaiTech University}
\icmlaffiliation{c}{Department of Statistics, Rutgers University}
\icmlaffiliation{d}{Department of Computer Science, University of North Carolina at Chapel Hill}
\icmlaffiliation{e}{College of Mathematics and Statistics, Chongqing University}

\icmlcorrespondingauthor{Hongxin Wei}{weihx@sustech.edu.cn}
\icmlcorrespondingauthor{Yue Qiu}{qiuyue@cqu.edu.cn}

\icmlkeywords{Machine Learning, ICML}

\vskip 0.3in
]




\printAffiliationsAndNotice{\icmlEqualContribution,\icmlWorkDone} 

\begin{abstract}
Conformal prediction is a statistical framework that generates prediction sets containing ground-truth labels with a desired coverage guarantee. 
The predicted probabilities produced by machine learning models are generally miscalibrated, leading to large prediction sets in conformal prediction.
 To address this issue, we propose a novel algorithm named \textit{Sorted Adaptive Prediction Sets} (SAPS), which discards all the probability values except for the maximum softmax probability. 
The key idea behind SAPS is to minimize the dependence of the non-conformity score on the probability values while retaining the uncertainty information.
In this manner, SAPS can produce compact prediction sets and communicate instance-wise uncertainty. 
Extensive experiments validate that SAPS not only lessens the prediction sets but also broadly enhances the conditional coverage rate of prediction sets.
\end{abstract}
\section{Introduction}
Machine learning is being deployed in many high-stakes tasks, such as autonomous driving \cite{bojarski2016end}, medical diagnostics \cite{caruana2015intelligible}, and financial decision-making. 
The trust and safety in these applications are critical, as any erroneous prediction can be costly and dangerous.
To assess the reliability of predictions, a popular solution is to quantify the model uncertainty, such as confidence calibration \citep{guo2017calibration}, MC-Dropout \citep{gal2016dropout}, and Bayesian neural network \citep{smith2013uncertainty}. 
However, these methods lack theoretical guarantees of the model performance. 
This gives rise to the importance of Conformal Prediction (CP) \citep{vovk2005algorithmic, shafer2008tutorial, balasubramanian2014conformal,angelopoulos2021gentle}, which yields prediction sets containing ground-truth labels with a desired coverage guarantee.

To determine the size of the prediction set, CP algorithms generally design non-conformity scores to quantify the deviation degree between a new instance and the training data. 
A higher non-conformity score is associated with a larger prediction set or region, indicating a lower confidence level in the prediction.
For example, Adaptive Prediction Sets (APS) \citep{romano2020classification} calculates the score by accumulating the sorted softmax values in descending order.
However, the softmax probabilities typically exhibit a long-tailed distribution, allowing for easy inclusion of those tail classes in the prediction sets.
To alleviate this issue, Regularized Adaptive Prediction Sets (RAPS) \citep{angelopoulos2020uncertainty} excludes unlikely classes by appending a penalty to classes beyond a specified threshold.
 Yet, the non-conformity score of RAPS involves unreliable softmax probabilities, leading to suboptimal performance in conformal prediction.
This motivates our question: \textit{does the probability value play a critical role in conformal prediction?}

In this work, we show that the value of softmax probability might be redundant information for conformal prediction. 
We provide an empirical analysis by removing the exact value of softmax probability while preserving the relative rankings of labels.
The results indicate that APS without probability value yields much more compact prediction sets than APS using the softmax outputs, at the same coverage rate.
Theoretically, we show that, by removing the probability value, the size of prediction sets generated by APS is consistent with the model prediction accuracy.
In other words, a model with higher accuracy can produce smaller prediction sets, using APS without access to the probability value. The details of the analysis are presented in Section~\ref{sec:motivation}.

Inspired by the analysis, our key idea is to minimize the dependence of the non-conformity score on the probability values, while retaining the uncertainty information. Specifically, we propose the \textit{Sorted Adaptive prediction sets} (dubbed \textbf{SAPS}), which discards all the probability values except for the maximum softmax probability in the construction of non-conformity score. After sorting in descending order, this can be achieved by replacing the non-maximum probability values with a constant. In effect, SAPS can not only produce sets of small size but also communicate instance-wise uncertainty. 
In addition, we show that as the constant approaches infinity, the expected value of set size from SAPS is asymptotically equivalent to that of APS without probability value (see Figure~\ref{fig:labmda_vs_size}).

To verify the effectiveness of our method, we conduct thorough empirical evaluations on common benchmarks, including CIFAR-10, CIFAR-100 \citep{krizhevsky2009learning}, and ImageNet \citep{deng2009imagenet}. The results demonstrate that SAPS is superior to the compared methods, including APS and RAPS. For example, our approach reduces the average size of prediction sets from 20.95 to 2.98 -- only $\frac{1}{7}$ of the prediction set size from APS. Compared to RAPS, we show that SAPS produces a higher conditional coverage rate and exhibits better adaptability to the instance difficulty. 
Code are publicly available at \href{https://github.com/ml-stat-Sustech/conformal_prediction_via_label_ranking}
{https://github.com/ml-stat-Sustech/conformal\_prediction\_via\_label\_ranking}.

 We summarize our contributions as follows:
 \begin{enumerate}
  \item We show that the probability value might be unnecessary for APS. Specifically, APS without probability value generates smaller prediction sets than vanilla APS. Theoretically, we show that this variant of APS can provide stable prediction sets, in which the set size is consistent with the prediction accuracy of models.

  \item We propose a novel non-conformity score -- SAPS that minimizes the dependency on probability value while retaining the uncertainty information. 
  
  \item  Extensive experimental results demonstrate the effectiveness of our proposed method. We show that SAPS lessens the prediction sets and broadly enhances the conditional coverage rate.
  
  \item We provide analyses to improve our understanding of the proposed method. In particular, we validate that both retaining the maximum softmax probability and confidence calibration are important for SAPS to generate compact prediction sets (see Section~\ref{sec:discussion}).
  
 \end{enumerate}

\section{Preliminaries }
In this work, we consider the multi-class classification task with $K$ classes.
Let $\mathcal{X}\subset\mathbb{R}^{d}$ be the input space and $\mathcal{Y}:=\{1,\dots,K\}$ be the label space.
We use $\hat{\pi}:\mathcal{X} \rightarrow \mathbb{R}^{K}$ to denote the pre-trained neural network that is used to predict the label of a test instance. Let $(X,Y)\sim \mathcal{P}_{\mathcal{X}\mathcal{Y}}$ denote a random data pair satisfying a joint distribution $\mathcal{P}_{\mathcal{X}\mathcal{Y}}$. Ideally, $\hat{\pi}_y(\boldsymbol{x})$ can be used to approximate the conditional probability of the class $y$ given an image feature $\boldsymbol{x}$, i.e., $\mathbb{P}[Y=y|X=\boldsymbol{x}]$. Then, the model prediction in classification tasks is generally made as: $\hat{y} = \underset{y\in\mathcal{Y}}{\arg\max} ~\hat{\pi}_y(\boldsymbol{x})$.



\paragraph{Conformal prediction.} To provide a formal guarantee for the model performance, conformal prediction \citep{vovk2005algorithmic} is designed to produce prediction sets containing ground-truth labels with a desired probability. Instead of predicting one-hot labels from the model outputs, the goal of conformal prediction is to construct a set-valued mapping $\mathcal{C}:\mathcal{X}\rightarrow 2^\mathcal{Y}$, which satisfies the \textit{marginal coverage}: 
\begin{equation}
\label{eq:validity}
    \mathbb{P}(Y\in \mathcal{C}(X))\geq 1-\alpha,  
\end{equation}
where $\alpha\in(0,1)$ denotes the desired error rate and $\mathcal{C}(X)$ is a subset of $\mathcal{Y}$.

Before deployment, conformal prediction begins with a calibration step, using the calibration set $\mathcal{D}_{cal}:=\{(\boldsymbol{x}_i,y_i)\}_{i=1}^n$. The samples of the calibration set are i.i.d., having been drawn from the distribution $\mathcal{P}_{\mathcal{X}\mathcal{Y}}$.
Specifically, we calculate a non-conformity score $s_i = S(\boldsymbol{x}_i,y_i)$ for each example $(\boldsymbol{x}_i,y_i)$ in the calibration set, where $s_i$ measures the degree of deviation between the given example and the training data. The $1-\alpha$ quantile of the non-conformity scores  $\{s_i\}_{i=1}^n$  is then determined as a threshold $\tau$. 
Formally, the value of $\tau$ can be obtained as shown below:
\begin{equation}\label{eq:threshold}\scalebox{1.1}{$
    \tau = \inf\{s:\frac{|\{i\in\{1,\dots,n\}: s_i\leq s   \}   |}{n}\geq \frac{\lceil (n+1)(1-\alpha)\rceil}{n}\}.$}
\end{equation}
During testing, we calculate the non-conformity score for each label given a new instance $\boldsymbol{x}_{n+1}$. Then, the corresponding prediction set $\mathcal{C}(\boldsymbol{x}_{n+1})$ comprises possible labels whose non-conformity score $S(\boldsymbol{x}_{n+1},y)$ falls within $\hat{\tau}$:
\begin{equation} \label{eq:cp_set}
   \mathcal{C}_{1-\alpha}(\boldsymbol{x}_{n+1};\tau) :=\{y\in\mathcal{Y}: S(\boldsymbol{x}_{n+1},y)\leq \tau \}.
\end{equation}

The above equation exhibits a nesting property of threshold, i.e., 
\begin{equation} \label{eq:nesting_property}
    \tau_1\leq \tau_2 \Longrightarrow  \mathcal{C}_{1-\alpha}(\boldsymbol{x}_{n+1};\tau_1) \subseteq \mathcal{C}_{1-\alpha}(\boldsymbol{x}_{n+1};\tau_2).
\end{equation}

This property shows that with a lower value of $\tau$, the model tends to produce a smaller prediction set, indicating a higher level of confidence in the prediction. Conversely, the increase of $\tau$ will enlarge the size of the prediction set, suggesting greater uncertainty of the prediction. In this manner, conformal prediction can be used to estimate the uncertainty or reliability of the model's predictions.

\begin{figure}[!t]
    \centering
    \resizebox{\linewidth}{!}{
    \begin{subfigure}{0.24\textwidth}
        \centering
        \raisebox{3.5mm}{\includegraphics[width=\textwidth]{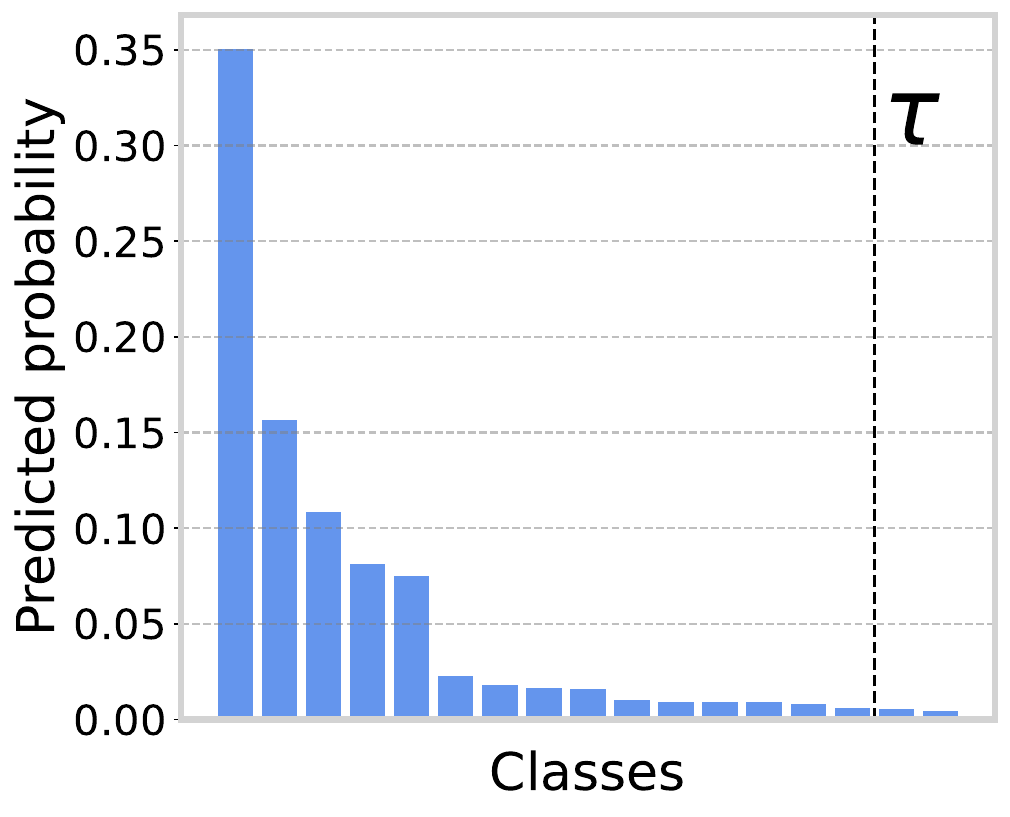}}
        \subcaption{}
        \label{fig:aps_example}
      \end{subfigure}
      \begin{subfigure}{0.23\textwidth}
        \centering
        \raisebox{-27mm}{\includegraphics[width=\linewidth]{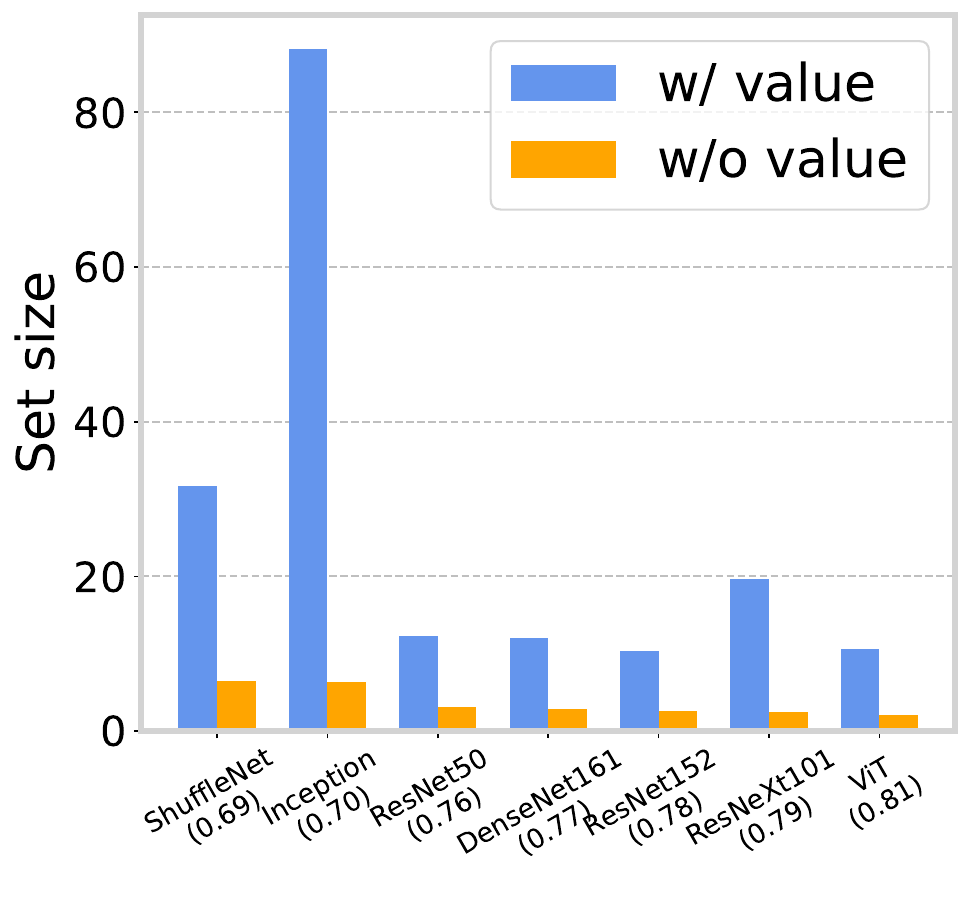}}
        \subcaption{}
        \label{fig:aps_wo_value}
      \end{subfigure}
      }
\caption{(a) Softmax probabilities for an instance from ImageNet are arranged in descending order. (b) Set size on various models. We use ``w/ value" and ``w/o value" to represent the vanilla APS and APS without probability values, respectively. The numbers in brackets represent the accuracy of the model. The sizes of the prediction sets are decreased after removing the probability value.}
  \end{figure}

\paragraph{Adaptive prediction sets (APS).}  In the APS method \citep{romano2020classification}, the non-conformity scores are calculated by accumulating the softmax probabilities in descending order. Formally, given a data pair $(\boldsymbol{x},y)$, the non-conformity score can be computed by:
\begin{equation}\label{eq:aps_score}\scalebox{0.95}{$
 S_{aps}(\boldsymbol{x},y,u;\hat{\pi}) := \sum\limits_{i=1}^{o(y,\hat{\pi}(\boldsymbol{x}))-1} \hat{\pi}_{(i)}(\boldsymbol{x})+u \cdot \hat{\pi}_{(o(y,\hat{\pi}(\boldsymbol{x})))}(\boldsymbol{x}),$}
\end{equation}

where $o(y,\hat{\pi}(\boldsymbol{x}))$ denotes the index of $\hat{\pi}_{y}(\boldsymbol{x})$ in the sorted softmax probabilities, i.e., $\hat{\pi}_{(1)}(\boldsymbol{x}),\dots,\hat{\pi}_{(K)}(\boldsymbol{x})$~, and $u$ is an independent random variable satisfying a uniform distribution on $[0,1]$. Similarly, the prediction set of APS with the error rate $\alpha$ for a test instance $\boldsymbol{x}_{n+1}$ is given by 
\begin{equation}\label{eq:aps_prediction_set}\scalebox{0.9}{$
    \mathcal{C}_{1-\alpha}(\boldsymbol{x}_{n+1},u_{n+1};\tau):=\{y\in\mathcal{Y}:S_{aps}(\boldsymbol{x}_{n+1},y,u_{n+1};\hat{\pi})\leq \tau \},$}
\end{equation}
where $u_{n+1}\sim U[0,1]$ and $\tau$ is defined by Equation~\ref{eq:threshold}. 
In particular, a larger value of $\tau$ will lead to a larger prediction set, representing higher uncertainty. Thus, the prediction set defined by Equation~\ref{eq:aps_prediction_set} satisfies the nesting property in Equation~\ref{eq:nesting_property}.

In addition, the prediction set $\mathcal{C}_{1-\alpha}(\boldsymbol{x}_{n+1},u_{n+1};\tau)$ with the $\tau$ obtained by Equation~\ref{eq:threshold} has a finite-sample marginal coverage guarantee, as stated in the following theorem.

\begin{theorem}\label{theorem:coverage_guarantee}
    \cite{angelopoulos2020uncertainty} Suppose the calibration data $(X_i,Y_i,U_i)_{i=1,\dots,n}$ and a test instance $(X_{n+1},Y_{n+1},U_{n+1})$ are exchangeable. Let the set-valued function $\mathcal{C}_{1-\alpha}(\boldsymbol{x},u;\tau)$ satisfy the nesting property of $\tau$ in Equation~\ref{eq:nesting_property}. For $\tau$ defined in Equation~\ref{eq:threshold}, we have the following coverage guarantee:
     $$P\left(Y_{n+1} \in \mathcal{C}_{1-\alpha} \left(X_{n+1}, U_{n+1};\tau\right)\right) \geq 1-\alpha.$$
\end{theorem}
In practice, the assumption of exchangeability in the above Theorem is promising as it is weaker than the i.i.d. assumption \cite{DBLP:conf/iclr/TengWZB0Y23}.
However, the softmax probabilities $\hat{\pi}(\boldsymbol{x})$ typically exhibit a long-tailed distribution, where the tail probabilities with small values can be easily included in the prediction sets. Consequently, APS tends to produce large prediction sets for all inputs, regardless of the instance difficulty. For example, in Figure~\ref{fig:aps_example}, the long-tail probability distribution results in the non-conformity scores of many classes falling within the threshold $\tau$. 
This motivates our analysis to investigate the role of probability values in conformal prediction.

\section{Motivation and method}
\subsection{Motivation}
\label{sec:motivation}
To analyze the role of probability values, we perform an ablation study by removing the influence of probability values in Equation~\ref{eq:aps_score}. In particular, after arranging softmax probabilities in descending order, we substitute them with a positive constant $\gamma$ (e.g., $\gamma=1$).

Formally, the modified non-conformity score for a data pair $(\boldsymbol{x},y)$ with a pre-trained model $\hat{\pi}$ can be give by:
  \begin{equation}\label{eq:constantaps_score}
  S_{cons}(\boldsymbol{x},y,u;\hat{\pi},\gamma) := \gamma \cdot \left[ o(y,\hat{\pi}(\boldsymbol{x}))-1+ u \right].
  \end{equation}
In what follows, we thoroughly probe the impacts of $\gamma$. We start by giving the following definition.

\begin{lemma}\label{definition:equivalent_scores}
    \cite{vovk2005algorithmic} Let $\mathcal{Z}$ be a space of examples. Given a significance level, two non-conformity score functions $A$ and $B$ could provide the same prediction sets if $A(z)\leq A(z^\prime) \Longleftrightarrow B(z) \leq B(z^\prime)$ for all $z,z^\prime \in \mathcal{Z}$. 
\end{lemma}

Lemma~\ref{definition:equivalent_scores} states that in conformal prediction, different non-conformity score functions are equivalent if they lead to the same non-conformity order for any data examples.
\begin{proposition}\label{proposition:aps_constant}
    For all $\gamma_1,\gamma_2>0$, the score functions $S_{cons}(\boldsymbol{x},y,u;\hat{\pi},\gamma_1)$ and $S_{cons}(\boldsymbol{x},y,u;\hat{\pi},\gamma_2)$ could provide the same prediction sets in conformal prediction.  
\end{proposition}
The proof of the above proposition is presented in Appendix~\ref{appendix:aps_constant}.
According to Proposition~\ref{proposition:aps_constant}, we can find that the value of $\gamma$ in $S_{cons}(\boldsymbol{x},y,u;\hat{\pi},\gamma)$ is irrelevant with the prediction sets. Thus, we fix the constant $\gamma$ to 1 for simplification. We conduct experiments on ImageNet \citep{deng2009imagenet} to compare the new non-conformity score to the vanilla APS. Here, we set the desired error rate as $10\%$, i.e., $\alpha=0.1$.
Following previous works \citep{romano2019conformalized,angelopoulos2020uncertainty,DBLP:conf/aaai/GhoshB0D23}, we first randomly split the test dataset of ImageNet into two subsets: a conformal calibration subset of size 30K and a test subset of size 20K. 
For the network architecture, we use seven models trained on ImageNet, with different levels of prediction performance (see Figure~\ref{fig:aps_wo_value}).
All models are calibrated by the temperature scaling procedure \citep{guo2017calibration} on the calibration dataset. Finally, experiments are repeated ten times and the median results are reported.

\paragraph{Probability values may not be necessary.} Figure~\ref{fig:aps_wo_value} presents the results on various models, using APS with/without the probability value. 
The results indicate that APS solely based on label ranking generates smaller prediction sets than the vanilla APS, across various models. 
For example, on the Inception model, removing the probability values reduces the set size from $88.18$ to $6.33$. Similarly, when applying a transformer-based ViT model \citep{touvron2021training}, APS without probability values also obtains a smaller set size. 
This comparison suggests that the probability values might be redundant for non-conformity scores in conformal prediction. We further explore this by theoretically analyzing the benefits of discarding softmax probability values in APS.

\paragraph{A theoretical interpretation.} The above empirical results demonstrate that the probability value is not a critical component of the non-conformity score for conformal prediction. 
Here, we provide a formal analysis of APS without probability value through the following theorem: 

 \begin{theorem}\label{theorem:constantaps}
 Let $A_r$ denote the accuracy of the top $r$ predictions for a trained model $\hat{\pi}$ on an infinite calibration set.
Given a significance level $\alpha$, there exists an integer $k$ satisfying  $A_{k}\geq 1-\alpha >A_{k-1}$. For any test instance $\boldsymbol{x} \sim \mathcal{P}_{\mathcal{X}}$ and an independent random variable $u\sim U[0,1]$, the size of the prediction set $\mathcal{C}_{1-\alpha}(\boldsymbol{x},u)$ generated by APS without probability value can be obtained by 
 \begin{align}\label{eq:constantaps_size}
 |\mathcal{C}_{1-\alpha}(\boldsymbol{x},u)|=
\left\{
\begin{aligned}
  &k,\quad\quad \mathrm{if } \hspace{1mm} u < \frac{1-\alpha - A_{k-1}}{A_{k}-A_{k-1}},  \\
  &k-1,\quad \mathrm{otherwise}.
\end{aligned}
\right.
 \end{align}
The expected value of the set size can be given by 
 \begin{equation}\label{eq:constantaps_expected_size}
  \mathbb{E}_{u \sim [0,1]}[|\mathcal{C}_{1-\alpha}(\boldsymbol{x},u)|] =  k-1+\frac{1-\alpha - A_{k-1}}{A_{k}-A_{k-1}}.
 \end{equation}
\end{theorem}
The proof of  Theorem~\ref{theorem:constantaps} can be found in Appendix~\ref{proof:theorem:constantaps}.
This is consistent with the theoretical results of RAPS, which indicates that the size of prediction sets generated by RAPS with heavy regularization will be at most equal to $k$~\cite{angelopoulos2020uncertainty}.
As indicated by Equation~\ref{eq:constantaps_expected_size}, the prediction set size generated by APS without probability value is consistent with $k$. 
In other words, a higher model accuracy will lead to a smaller value of $k$, indicating smaller prediction sets. 
This argument is clearly supported by experimental results shown in Figure~\ref{fig:aps_wo_value}. 
In particular, we observe that using APS without probability value, models with higher accuracy produce smaller prediction sets, while the vanilla APS does not exhibit this characteristic. 
For example, the vanilla APS yields larger prediction sets for ResNeXt101 than for ResNet152, despite ResNeXt101 achieving higher predictive accuracy than ResNet152. The analysis demonstrates the advantage of removing probability value in APS, via decreasing the sensitivity to tail probabilities.


\subsection{Method}

In the above analysis, we demonstrate that removing the probability values in APS can largely decrease the size of prediction sets.
Yet, the expected value of the set size (shown in Equation~\ref{eq:constantaps_size}) will oscillate between $k-1$ and $k$, using APS without the probability value. This implies a shortcoming of the modified non-conformity score in adaptation to instance-wise uncertainty, leading to redundant prediction sets for easy examples.



To alleviate this limitation, we propose a novel conformal prediction algorithm, named \textit{Sorted Adaptive Prediction Sets} (SAPS). The key idea behind SAPS is to minimize the dependence of the non-conformity score on the probability values, while retaining the uncertainty information. In particular, we discard all the probability values except for the maximum softmax probability (MSP), which is usually used to measure the model confidence in the prediction \cite{hendrycks2016baseline}. 

Formally, the non-conformity score of SAPS for a data pair $(\boldsymbol{x},y)$ can be calculated as 
  \begin{equation}\label{eq:SAPS_score}\scalebox{0.85}{$
     S_{saps}(\boldsymbol{x},y,u;\hat{\pi}) := \left\{ 
      \begin{array}{ll}
        u \cdot\hat{\pi}_{max} (\boldsymbol{x}), \qquad \text{if\quad} o(y,\hat{\pi}(\boldsymbol{x}))=1,\\
        \hat{\pi}_{max} (\boldsymbol{x}) +(o(y,\hat{\pi}(\boldsymbol{x}))-2+u) \cdot \lambda, \quad \text{else},
      \end{array}
    \right.$}
 \end{equation} 
 where $\lambda$ is a hyperparameter representing the weight of ranking information,  $\hat{\pi}_{max} (\boldsymbol{x})$ denotes the maximum softmax probability and $u$ is a uniform random variable. 

During inference, we calculate the non-conformity score for each label. In particular, the score is equivalent to that of APS for the label with the maximum softmax probability, while the scores of the resting labels are composed of the MSP and the ranking information from 2 to $o(y,\hat{\pi}(\boldsymbol{x}))$, i.e., $(o(y,\hat{\pi}(\boldsymbol{x}))-2+u)$. In this way, the new score of SAPS retains instance-wise uncertainty and mitigates the undesirable influence of tail probabilities.

 With the new non-conformity score, the prediction sets of SAPS can be established according to Equation~\ref{eq:aps_prediction_set}, following APS and RAPS. Consequently, we can conclude that SAPS also satisfies the marginal coverage guarantee, as stated in Theorem~\ref{theorem:coverage_guarantee}. 
Moreover, We provide an ablation experiment to investigate the relation between the prediction sets and the ranking weight $\lambda$ in Section~\ref{sec:discussion}. 

\begin{table*}[!t] 
\centering 
\caption{Performance comparison of various methods with different error rates $\alpha$. We conduct experiments on ImageNet and CIFAR-10/100 datasets. \textbf{Bold} numbers indicate optimal performance.} 
\label{table:average_size} 
\renewcommand\arraystretch{1.2}
\resizebox{\textwidth}{!}{
\setlength{\tabcolsep}{2mm}{ 
\begin{tabular}{lcccccccccccc } 
\toprule 
 & \multicolumn{6}{c}{$\alpha=0.1$}  & \multicolumn{6}{c}{$\alpha=0.05$} \\ 
\cmidrule(r){2-7}  \cmidrule(r){8-13}  
 & \multicolumn{3}{c}{Coverage}  & \multicolumn{3}{c}{Size $\downarrow$} & \multicolumn{3}{c}{Coverage}  & \multicolumn{3}{c}{Size $\downarrow$}\\ 
\cmidrule(r){2-4}  \cmidrule(r){5-7} \cmidrule(r){8-10}  \cmidrule(r){11-13}  
Datasets   & APS & RAPS &SAPS & APS & RAPS & SAPS & APS & RAPS & SAPS & APS & RAPS & SAPS\\ 
\midrule 
 ImageNet& 0.899 & 0.900 & 0.900 & 20.95 & 3.29 & \textbf{2.98} & 0.949 & 0.950 & 0.950 & 44.67 & 8.57 & \textbf{7.55}  \\ 
 CIFAR-100& 0.899 & 0.900 & 0.899 & 7.88 & 2.99 & \textbf{2.67} & 0.950 & 0.949 & 0.949 & 13.74 & 6.42 & \textbf{5.53} \\ 
 CIFAR-10& 0.899 & 0.900 & 0.898 & 1.97 & 1.79 & \textbf{1.63} & 0.950 & 0.950 & 0.950 & 2.54 & 2.39 & \textbf{2.25}  \\ 
\bottomrule 
\end{tabular}}} 
\end{table*} 


\section{Experiments}
\label{sec:exp}

\subsection{Experimental Setup}
\label{sec:experimental_setup}
\paragraph{Datasets.} We consider three prominent datasets in our experiments: ImageNet \citep{deng2009imagenet}, CIFAR-100  and CIFAR-10 \citep{krizhevsky2009learning}, which are common benchmarks for conformal prediction. In particular, for ImageNet, its test dataset of 50,000 images is divided, allocating 30,000 images to the calibration set and 20,000 images to the test set. For both CIFAR-100 and CIFAR-10, the associated test dataset of 10,000 images is uniformly divided into two subsets: a calibration set and a test set, each comprising 5,000 images.

\paragraph{Models.} We employ twelve different classifiers, including nine standard classifiers, two transformer-based models, i.e., ViT \citep{dosovitskiy2020image} and DeiT \citep{touvron2021training}, and a Vision-Language Model named CLIP \citep{radford2021learning}. Aside from CLIP with zero-shot prediction capabilities,  the remaining models are the models pre-trained on ImageNet. For  CIFAR-10 and CIFAR-100, these models will be fine-tuned on the pre-trained models. Moreover, all classifiers are calibrated by the Temperature scaling procedure \citep{guo2017calibration} before applying CP methods.

\paragraph{Conformal prediction algorithms.}We compare the proposed method against APS \citep{romano2020classification} and RAPS \citep{angelopoulos2020uncertainty}. For methods that has hyper-parameters, we choose the hyper-parameter that achieves the smallest set size on a validation set, which is a subset of the calibration set. Specifically, we tune the regularization hyperparameter of RAPS in $\{0.001,0.01,0.1,0.15,\dots,0.5\}$  and hyperparameter $\lambda$ in $\{0.02,0.05,0.1,0.15,\dots,0.6\}$ for SAPS.
 All experiments are conducted with ten trials, and the median results are reported.
 
 Additionally, we provide the details about the calibration process of conformal prediction algorithms, as shown:
 \vspace{-5pt}
 \begin{enumerate}[itemsep=-1pt]
     \item \textbf{Split:} we split the full calibration set into a validation set and a calibration set ($20:80$ in this work);
     \item \textbf{Tuning:} we use the validation set to choose proper hyperparameters of non-conformity score function;
    \item \textbf{Conformal Calibration:} we use the calibration set to calculate the threshold $\tau$.
 \end{enumerate}
  \vspace{-5pt}
For CP methods with hyperparameters (i.e., RAPS and SAPS), we use the "Split" and "Tuning" steps to tune the hyperparameters. For APS, we use the full calibration set. Therefore, all methods have access to the same dataset for fair comparison.

\begin{figure*}[!t]
    \centering
    \resizebox{17cm}{!}{
    \begin{subfigure}{0.33\textwidth}
        \centering
        \includegraphics[width=\linewidth]{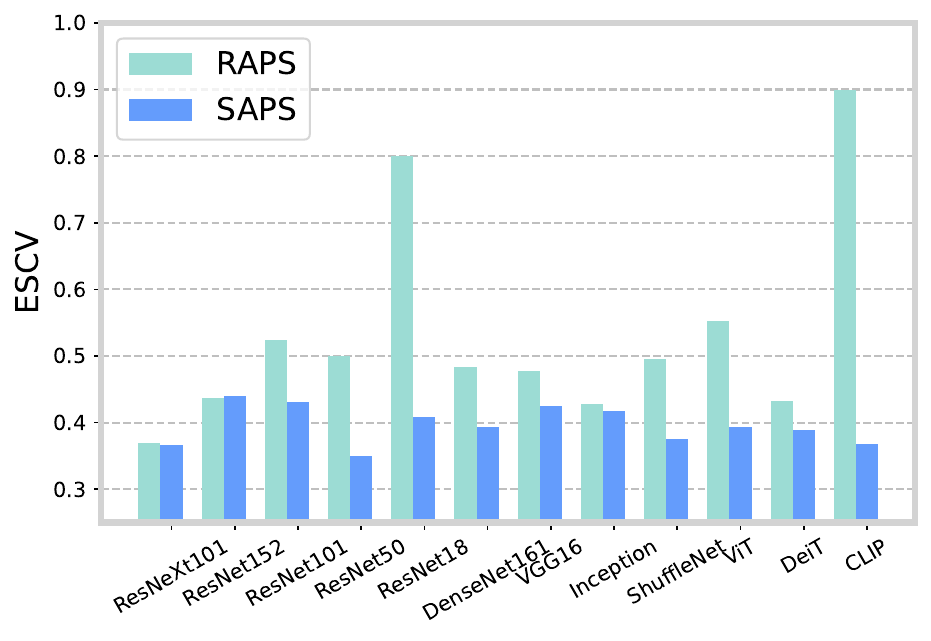}
        \subcaption{}
        \label{fig:escv_imagenet}
      \end{subfigure}
        \hspace{0.2cm}
      
        \hspace{0.2cm}
      \begin{subfigure}{0.33\textwidth}
        \centering
        \includegraphics[width=\linewidth]{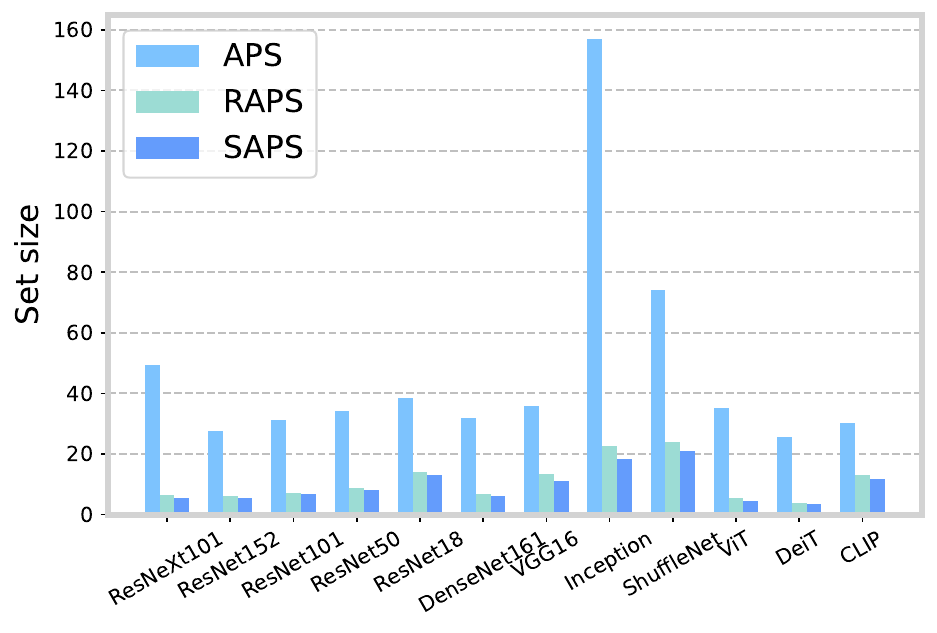}
        \subcaption{}
        \label{fig:size_imagenetv2}
      \end{subfigure}
      \begin{subfigure}{0.33\textwidth}
        \centering
        \includegraphics[width=\linewidth]{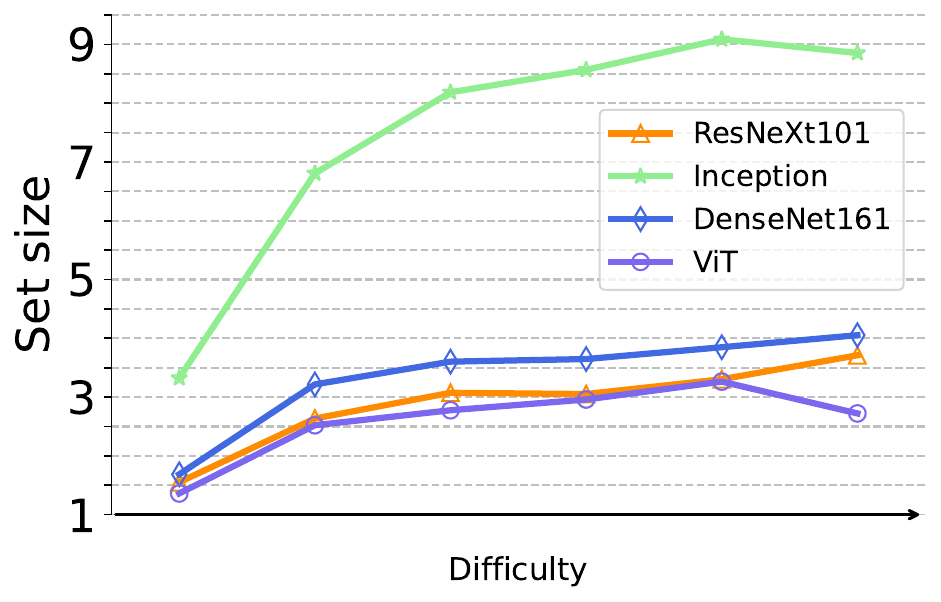}
        \subcaption{}
        \label{fig:size_vs_diff}
      \end{subfigure}
      }
      \caption{ (a) ESCV  with different models on ImageNet with $\alpha=0.1$.  A good conformal prediction algorithm should keep the y-axis (e.g., ESCV) small. The results show that SAPS outperforms RAPS on most cases. (b) Set size on ImageNet-V2 at $\alpha=0.1$. (c) Set size of various difficulties for multiple models on ImageNet. Small sets are required for easy examples, while hard ones require large sets.}    \label{fig:ESCV}

  \end{figure*}

\textbf{Evaluation.} The primary metrics used for the evaluation of prediction sets are set size (average length of prediction sets, small value means high efficiency) and marginal coverage rate (fraction of testing examples for which prediction sets contain the ground-truth labels).  These two metrics can be formally represented as :
$$\operatorname{Size}  = \frac{1}{N_{test}}\sum_{i=1}^{N_{test}} |\mathcal{C}(\boldsymbol{x}_i)|$$
$$\operatorname{Coverage\ rate} = \frac{1}{N_{test}}\sum_{i=1}^{N_{test}}\mathbf{1}(y_i\in\mathcal{C}(\boldsymbol{x}_i))$$

\emph{Conditional coverage rate.}  
In this work, we propose an alternative metric to the \emph{Size-stratified Coverage Violation} \citep{angelopoulos2020uncertainty} named \emph{Each-Size Coverage Violation} (ESCV), which is  given by:
\begin{equation*}
  \scalebox{0.95}{$\operatorname{ESCV}(\mathcal{C},K) = \sup\limits_{j}{ \max( {0,1-\alpha-\frac{  \left| \{i \in \mathcal{J}_{j} : y_{i} \in \mathcal{C} \left(\boldsymbol{x}_{i}\right) \} \right|}{\left|\mathcal{J}_{j}\right|}})}$}
\end{equation*}
where $\mathcal{J}_{j} = \{i: | \mathcal{C}\left(\boldsymbol{x}_{i}\right)|= j\}$ and $j\in\{1,\dots,K\}$.
Specifically, ESCV measures the most significant violation of prediction sets. This metric is practical because it only requires the set size, and is suitable for any classification problem, spanning from binary classes to large classes.

\subsection{Results}
\label{sec:results}

\textbf{SAPS generates smaller prediction sets.} In Table~\ref{table:average_size}, the performance of set sizes and coverage rates for various classification tasks are presented. We can observe that the coverage rate of all conformal prediction methods is close to the desired coverage $1-\alpha$.
At different significance levels (i.e., $0.1$ and $0.05$), the prediction set size is consistently reduced by SAPS for ImageNet, CIFAR-100 and CIFAR-10, compared to APS and RAPS.
For example, when evaluated on ImageNet, SAPS reduces the average set size from $20.95$ of APS to $2.98$. 
Moreover, as the scale of the classification task increases, the efficiency improvement achieved by SAPS becomes increasingly evident. Overall, the experiments show that our method has the desired coverage rate and a smaller set size than APS and RAPS.
Due to space constraints, we only report the average results of multiple models on various classification tasks in Table~\ref{table:average_size}, and detailed results for each model are available in Appendix~\ref{appendix:set_size}.

\textbf{Experiments on distribution shifts.}  We also verify the effectiveness of our method on the new distribution, which is different from the training data distribution. Specifically, we divide the test dataset of \mbox{ImageNet-V2}~\citep{recht2019imagenet}, which exhibits a distribution shift compared to the ImageNet, equally into a calibration set containing  5000 images and a test set containing 5000 images.  Then, the test models are only pre-trained on ImageNet and not be fine-tuned. 
As shown in Figure~\ref{fig:size_imagenetv2}, the result shows that under $\alpha=0.1$, our method can also generate the smallest sets when the conformal calibration set and the test set come from a new distribution.

\textbf{SAPS acquires lower conditional coverage violation.}  
 In Figure~\ref{fig:escv_imagenet}, we demonstrate that SAPS not only outperforms in efficiency but also boosts the conditional coverage rate, i.e., ESCV.  
 Given that our study primarily focuses on improving the efficiency of prediction sets, the comparison of ESCV is exclusively conducted between SAPS and RAPS.
 The results, shown in Figure~\ref{fig:escv_imagenet}, demonstrate that on ImageNet, SAPS would get smaller ESCV than RAPS for most models. For example, on  CLIP,  SAPS reduces the ESCV from $0.9$ to $0.37$. In addition, on ImageNet, we can observe that the ESCV of SAPS  for different models is more stable than RAPS. Specifically, the ESCV of SAPS can keep a low value on most models, but in the case of RAPS, the maximum ESCV even gets $0.9$.  The detailed results on CIFAR-10 and CIFAR-100 are provided in Appendix~\ref{appendix:escv}. 

 Moreover, to further confirm that SAPS achieves superior conditional coverage than RAPS, we report the results of \textit{Size-stratified Coverage Violation} (SSCV). In this experiment, the coarse partitioning of the set sizes for SSCV is set as: 0-1, 2-3, 4-10, 11-100, and 101-1000, which follows the experimental setting of RAPS~\cite{angelopoulos2020uncertainty}. The average SSCV of various models on three datasets are shown in Table~\ref{table:sscv}, with the experimental setup for ImageNet-V2 being consistent with the description provided in the preceding paragraph.
 We can observe that SAPS outperforms RAPS in conditional coverage. For instance, on ImageNet, SAPS records an SSCV of $0.23$, which is $0.05$ lower than that of RAPS. Although SAPS has better conditional coverage than RAPS, note that SSCV calculates the worst conditional coverage over the human-defined enclosures of set sizes, where different partitions lead to inconsistent SSCV results. Consequently, avoiding the human factor may lead to fair comparison, such as ESCV.

 \begin{table}[!t] 
\centering 
\caption{The average SSCV and ESCV of various models on three datasets. Bold numbers indicate optimal performance.}
\label{table:sscv} 
\begin{tabular}{cccc}
\toprule 
& ImageNet & ImageNet-V2 & CIFAR100 \\
\cmidrule(r){2-4}
Metrics & \multicolumn{3}{c}{RAPS/SAPS(ours)}  \\ 
\midrule 
SSCV$\downarrow$ & 0.28/\textbf{0.23} & 0.14/\textbf{0.13} & 0.21/\textbf{0.18} \\ 
ESCV$\downarrow$ & 0.53/\textbf{0.40} & 0.43/\textbf{0.41} & 0.37/\textbf{0.30} \\ 
\bottomrule 
\end{tabular}
\end{table}

\section{Discussion}
\label{sec:discussion}

\begin{table*}[!t] 
\centering 
\caption{Set size and ESCV for RAPS ($k_{r}=1$) and SAPS. We report the average value across various models with $\alpha=0.1$. The detailed results of each model are provided in the Appendix~\ref{appendix:RAPSK1}. \textbf{Bold} numbers indicate optimal performance.}
\label{table:RAPSK1} 
\renewcommand\arraystretch{1.2}
\begin{tabular}{lcccccc} 
\toprule 
 & \multicolumn{2}{c}{Coverage}  & \multicolumn{2}{c}{Size $\downarrow$} & \multicolumn{2}{c}{ESCV $\downarrow$} \\ 
\cmidrule(r){2-3}  \cmidrule(r){4-5} \cmidrule(r){6-7} 
Datasets   & RAPS($k_{r}=1$) & SAPS  & RAPS($k_{r}=1$) & SAPS  & RAPS($k_{r}=1$) & SAPS\\ 
\midrule 
 ImageNet&0.900 & 0.900 & 3.24 & \textbf{2.98} & 0.631 & \textbf{0.396}  \\ 
 CIFAR-100&0.899 & 0.899 & 2.79 & \textbf{2.67} & 0.390 & \textbf{0.302}  \\ 
 CIFAR-10&0.900 & 0.898& \textbf{1.62} & 1.63 & 0.138 & \textbf{0.089}  \\ 
\bottomrule 
\end{tabular}
\end{table*}

\begin{figure*}[!t]
    \centering
      \begin{subfigure}{0.24\linewidth}
        \includegraphics[width=\linewidth]{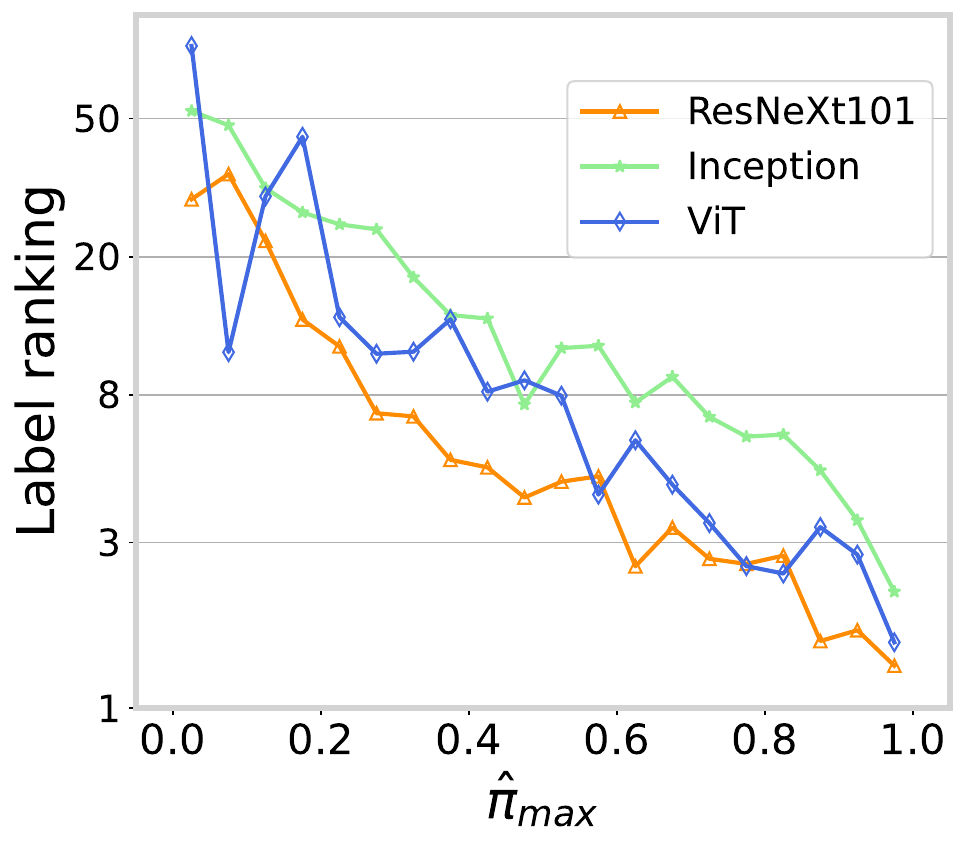}
        \caption{}
        \label{fig:confidences_vs_label_ranking}
      \end{subfigure}
    \begin{subfigure}{0.24\textwidth}
        \centering
        \includegraphics[width=\linewidth]{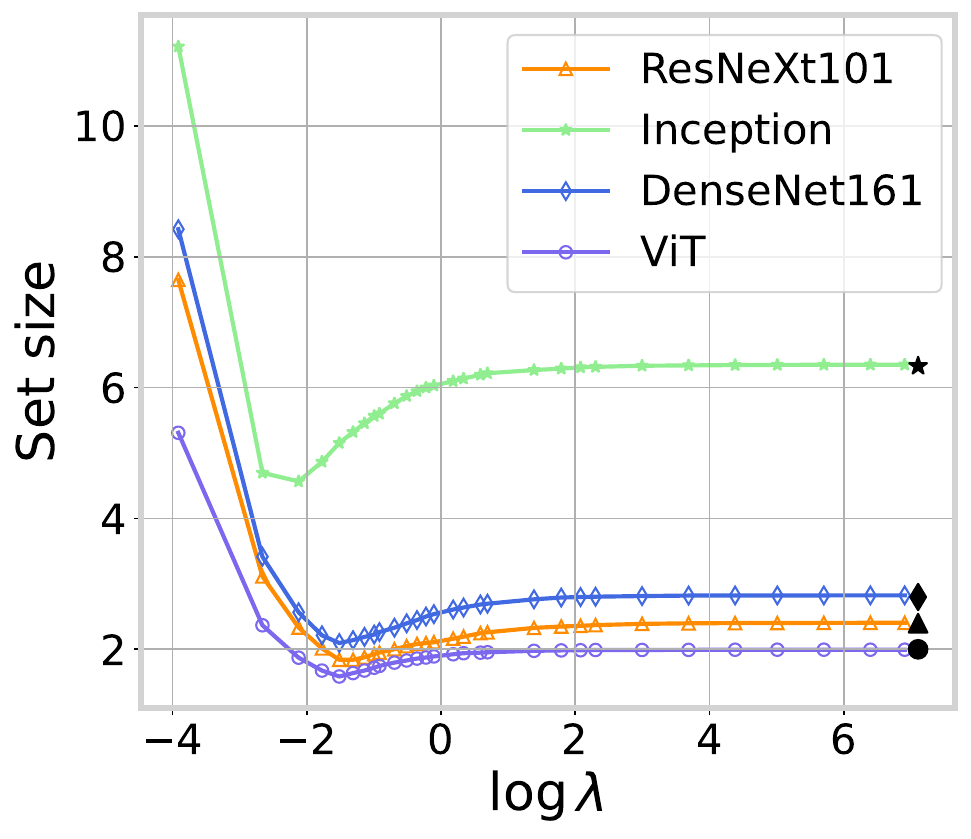}
        \caption{}
        \label{fig:labmda_vs_size}
    \end{subfigure}
    \begin{subfigure}{0.24\textwidth}
        \centering
        \includegraphics[width=\linewidth]{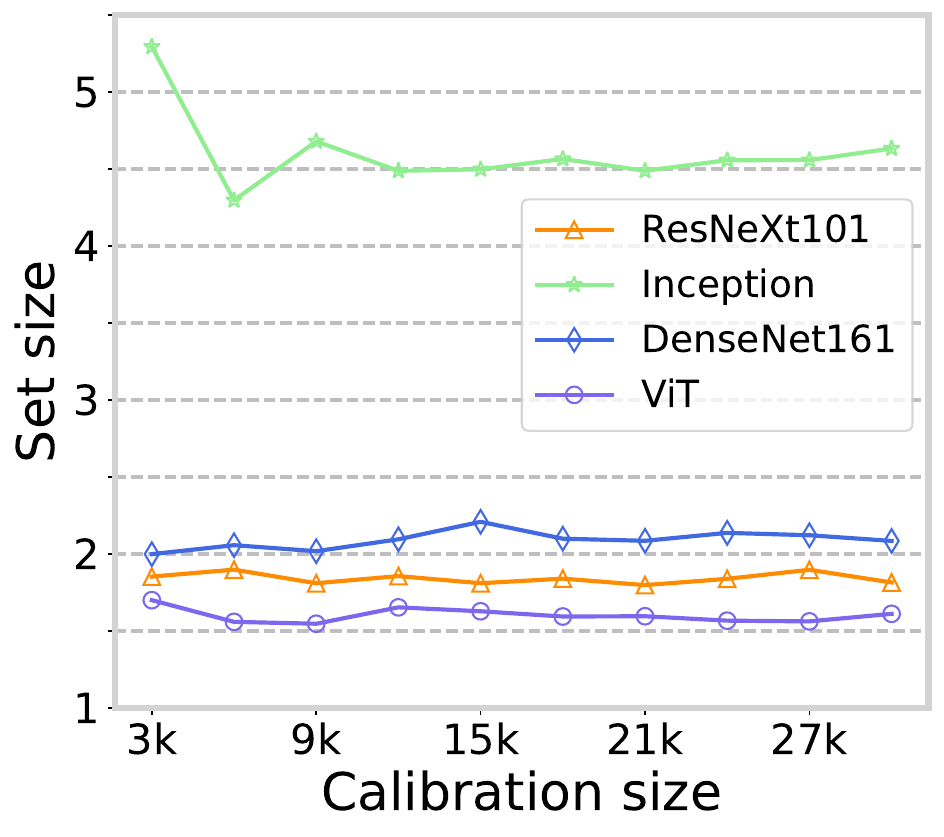}
        \caption{}
        \label{fig:number_vs_size}
    \end{subfigure}
    \begin{subfigure}{0.24\textwidth}
        \centering
        \includegraphics[width=\linewidth]{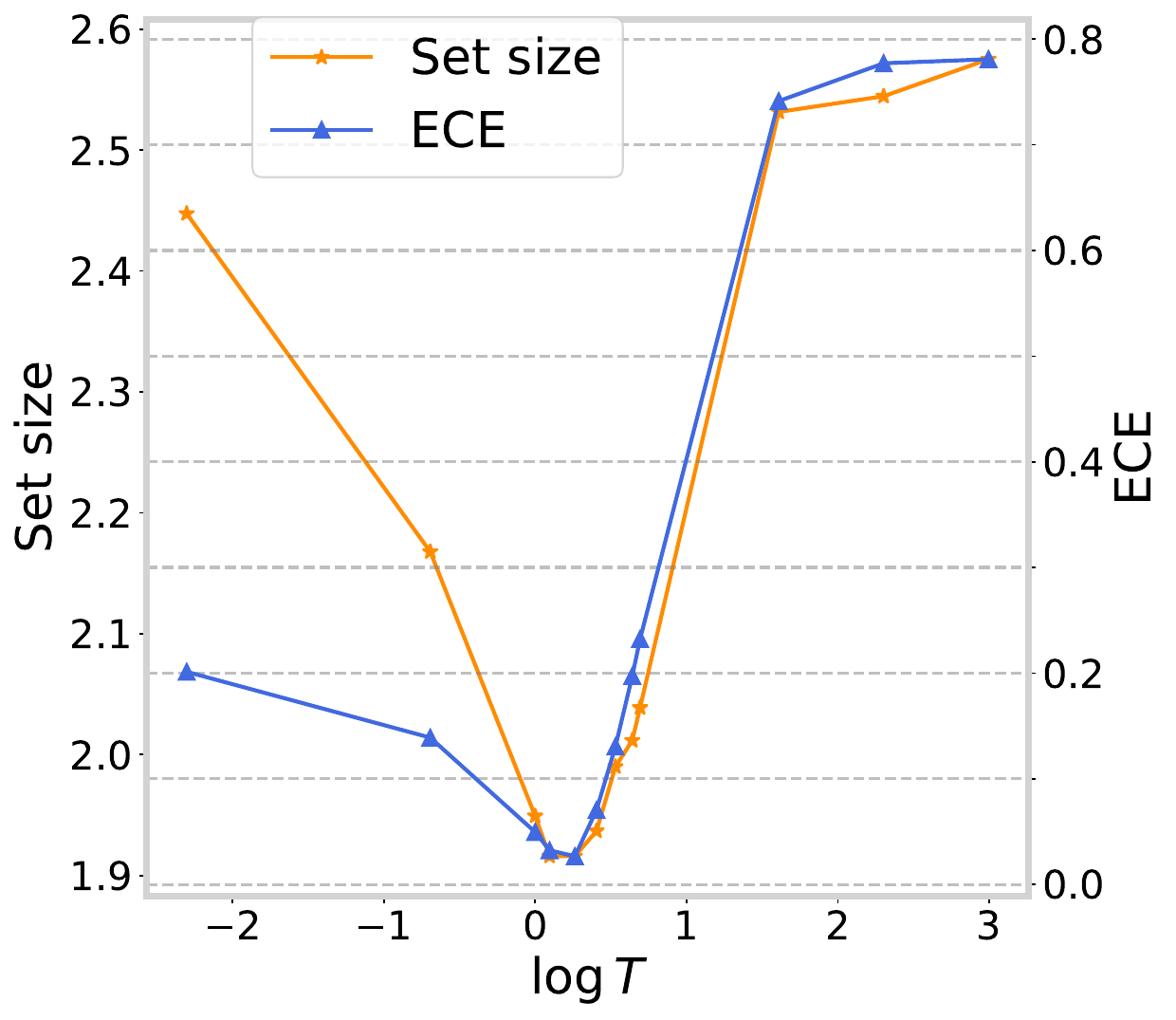}
        \caption{}
        \label{fig:ece_vs_size}
    \end{subfigure}
    \caption{(a) The average ground-truth label ranking under different maximum softmax probabilities. Higher $\hat{\pi}_{max}$ have smaller label ranking. (b)  Effect of the $\lambda$ on set size across various models. The black markers ({\footnotesize$\bigstar,\blacklozenge,\blacktriangle,\bullet$})  represent the results of APS without probability value. (c)  Effect of the calibration dataset size on set size across various models. (d) Relationship between temperature and the set size of  SAPS on ResNet152, where the horizon axis represents the log transformation of temperature $T$.}
    \label{fig:analysis}
\end{figure*}

\paragraph{SAPS exhibits adaptation.}
In the literature of conformal prediction, the size of prediction sets is expected to represent the inherent uncertainty of the classifier's predictions. Specifically, prediction sets should be larger for hard examples than for easy ones \cite{angelopoulos2021gentle,angelopoulos2020uncertainty}. 
In this work, we employ the rank of the ground-truth labels in the sorted softmax probabilities to denote the difficulty. For instance, examples with serious difficulty are assigned high ranks for their ground-truth labels. For simplicity, we choose a relatively coarse partitioning of image difficulty: 1, 2-3, 4-10, 11-100, and 101-1000. Then, we conduct an experiment on ImageNet to analyze the set size of different difficulties for SAPS.

In Figure~\ref{fig:size_vs_diff}, a salient observation is that the set size rises with increasing difficulties for different models. 
For example, on the Inception model, SAPS enlarges the length of the prediction sets from approximately $3.5$ for the simplest examples to nearly $9$ for the hardest examples. Moreover, we notice that the classifier with a higher accuracy rate can have a smaller set size under different levels of difficulty compared to the classifier with a lower accuracy rate. For example, the ResNeXt101 model has a smaller set size than DenseNet161 across different difficulties.
\vspace{-10pt}

\begin{table}[!t] 
\centering 
\caption{Set size about retained top-$k$ probability values on the ImageNet. We report the average set size across various models with $\alpha= 0.1$. \textbf{Bold} numbers indicate optimal performance.}
\label{table:retianed_probability} 
\begin{tabular}{lccccc} 
\toprule 
 & \multicolumn{5}{c}{$k$}  \\
 \cmidrule(r){2-6} 
Model &1 &2 &3 &4 &5 \\
\midrule 
 ResNeXt101 & \textbf{1.80} & 1.94  & 2.03 & 2.23 & 2.42   \\ 
 DenseNet161 & \textbf{2.12} & 2.16  & 2.27  & 2.46  & 2.65    \\ 
 ViT &\textbf{1.60}  & 1.67 & 1.76 & 1.86 & 1.97   \\ 
 Inception & 4.60  & \textbf{4.50}  & 4.55  & 4.77  & 4.95    \\ 
\bottomrule 
\end{tabular}
\end{table}

\paragraph{Retained probability values.} To mitigate the redundancy of prediction sets for easy examples, we retain MSP in the softmax probabilities, as shown in Equation~\ref{eq:SAPS_score}. In this manner, the SAPS method can not only produce efficient prediction sets but also communicate instance-wise uncertainty, i.e., adaptation.  
To further understand why retaining MSP in SAPS can decrease the set size of easy examples, we present the ground-truth label ranking of different MSP for three different models on ImageNet in Figure~\ref{fig:confidences_vs_label_ranking}. We can observe that MSP is negatively related to ground-truth label ranking. Therefore, the ground-truth labels of easy examples (high MSP values) are generally highly ranked, which are expected to give compact prediction sets. In other words, the maximum softmax probability of a sample contains the information of its difficulty, allowing SAPS to produce adaptive prediction sets.

Here, we further discuss the effect of other softmax probability values on set size.  Specifically, we retain the first $k$ maximum softmax probability values, where $k\in\{1,2,3,4,5\}$~. We conduct experiments on the ImageNet with four classifiers of different architectures. In Figure~\ref{fig:confidences_vs_label_ranking}, we show that retaining MSP achieves the best performance on most models. Additionally, more unreliable probability values involved in the score function will lead to larger prediction sets. 
Overall, retaining MSP in SAPS is the key to decreasing set size and preserving uncertainty. 

To explore the sufficiency of MSP for conformal prediction, we provide an ablation study with some variants of SAPS by maintaining more probability values. Specifically, we compare five variants that preserve the top $k\in \{1, 2, 3, 4, 5\}$ values of softmax probabilities, respectively. The results of ImageNet using the four classifiers are reported in Table~\ref{table:retianed_probability} of the manuscript. The results show that the vanilla SAPS with only MSP can achieve the best performance in most cases. For instance, with the ResNeXt101 model, only keeping the maximum softmax probability boosts the set size of $k=5$ from $2.42$ to $1.80$, a $0.62$ of direct improvement.
In addition, retaining more tail-end probability values in the score function leads to larger prediction sets. Overall, retaining MSP is sufficient for SAPS.

\paragraph{Effect of $\lambda$ and the calibration size.} In SAPS, we choose an optimal $\lambda$ by performing a search over a sequence to minimize the set size on a validation set. In this work, the validation set constitutes 20\% of the calibration set. Here, we provide an experimental procedure to show \textit{whether set size is sensitive to $\lambda$ and calibration size.}
To this end, we conduct two experiments on ImageNet to analyze the effects of $\lambda$ and the size of the calibration set.  

We present the results of four models in Figure~\ref{fig:analysis}. Indeed,
Figure~\ref{fig:labmda_vs_size} illustrates that one can efficiently utilize grid search to find the optimal $\lambda$. Furthermore, as depicted in Figure~\ref{fig:number_vs_size}, nearly all models maintain stable results when the number of calibration sets increases.
Overall, the results demonstrate that the set size is not sensitive to variations in $\lambda$ and the calibration size.

\paragraph{SAPS vs. RAPS ($k_{r}=1$).} While SAPS has demonstrated strong promise,  it shares a similarity in the definition of non-conformity scores  with RAPS ($k_{r}=1$), as shown below:

\begin{align*}
    S_{raps}(\boldsymbol{x},y,u,\hat{\pi}) &=  \sum_{i=1}^{o(y,\hat{\pi}(\boldsymbol{x}))-1} \hat{\pi}_{(i)}(\boldsymbol{x}) + u *\hat{\pi}_{o(y,\hat{\pi}(\boldsymbol{x}))}(\boldsymbol{x}) \\
    &+ \phi \cdot (o(y,\hat{\pi}(\boldsymbol{x}))-k_{r})^{+}.
 \end{align*}
 
 Here, $\phi$ represents the weight of regularization and $(z)^{+}$ denotes the positive part of $z$. To this end, we conduct experiments on  CIFAR-10, CIFAR-100, and ImageNet to compare SAPS and RAPS ($k_{r}=1$), with $\alpha=0.1$.
 
 As indicated in Table~\ref{table:RAPSK1}, SAPS outperforms RAPS ($k_{r}=1$) in large-scale classification scenarios, achieving smaller prediction sets and lower conditional coverage violations.  In the small-scale classification task (i.e., CIFAR-10), SAPS produces a comparable set size with RAPS ($k_{r}=1$), and the ESCV of SAPS was more than 1.5 times as small as those from RAPS.  Overall, employing a constant to substitute the noisy probabilities is an effective way to alleviate the negative implications of noisy probabilities further.
 
\paragraph{Relation to temperature scaling.} In the literature, temperature scaling calibrates softmax probabilities output of models by minimizing Expected Calibration Error (ECE), leading to a reliable maximum probability \cite{guo2017calibration}. Moreover, as defined in Equation~\ref{eq:SAPS_score}, the remaining probability value used in the non-conformity score function is the maximum probability. Thus, a question arises: \textit{what is a relation between temperature scaling and the set size of SAPS?} Here, we vary the value of temperature $T=~\{0.1, 0.5, 1, 1.1, 1.3, 1.5, 1.7, 1.9, 2, 5, 10, 20\}$ in temperature scaling. We utilize SAPS to test the ResNet152 model calibrated by different temperatures on the ImageNet benchmark. 
The results indicate that there exists a consistency between the temperature value and the set size.

As illustrated in Figure~\ref{fig:ece_vs_size}, the optimal temperature on ECE, i.e., $1.3$coincides with the temperature that yields the smallest set size. 
Moreover, we can observe that an increase in ECE results in an expanded set size.
Indeed, temperature scaling can not change the permutation of the softmax probabilities but improves the confidence level of the maximum probability, resulting in the non-conformity scores of SAPS being more reliable.
Overall, for SAPS, confidence calibration is important to generate compact prediction sets.

\section{Related work}
Conformal prediction (CP) is a statistical framework characterized by a finite-sample coverage guarantee~\citep{vovk2005algorithmic}.  
It has been utilized in various tasks including regression~ \citep{lei2014distribution,romano2019conformalized}, classification~ \citep{sadinle2019least},  structured prediction~ \citep{bates2021distribution}, Large-Language Model~ \citep{kumar2023conformal,ren2023robots,quach2023conformal,su2024api}, Diffusion Model~\citep{teneggi2023trust, horwitz2022conffusion}, robots control~\citep{wang2023conformal}, graph neural network~\citep{wijegunawardana2020node, clarkson2023distribution,song2024similarity} and so on. 
Conformal prediction also is deployed in other applications, such as human-in-the-loop decision making  \citep{straitouri2022improving,cresswell2024conformal}, automated vehicles~\citep{bang2024safe}, and scientific machine learning~\citep{moya2024conformalized,podina2024conformalized}.
Besides the inductive conformal prediction~\cite{vovk2005algorithmic},   most CP methods follow an inductive conformal framework (or split conformal prediction) with a held-out calibration set to cope with the relative computational inefficiency of conformal predictors~\citep{10.1007/3-540-36755-1_29,Papadopoulos2002QualifiedPF, 1181895}, as in the main paper \cite{lei2015conformal}.
Additionally, there are various variants of conformal predictors based on cross-validation \citep{vovk2015cross} or jackknife (i.e., leave-one-out) \citep{barber2021predictive}.
The primary focal points of CP are reducing prediction set size and enhancing coverage rate.  

\paragraph{Inefficiency.} Strategies to improve the efficiency of prediction sets can be roughly split into the following two branches. 
The first avenue of research focuses on developing new training algorithms to reduce the average prediction set size  \citep{bellotti2021optimized,colombo2020training,chen2021learning,DBLP:conf/iclr/StutzDCD22,einbinder2022training,bai2022efficient,fisch2021few,yang2021finite}. Those training methods are usually computationally expensive due to the model retraining and the high complexity of optimization. Thus, CP has also been studied in the context of Early Stopping \cite{DBLP:conf/icml/LiangZS23}, allowing model selection leading to compact prediction sets while keeping coverage guarantees.
The other avenue involves leveraging post-hoc technologies, such as novel score functions~\citep{romano2020classification, angelopoulos2020uncertainty,DBLP:conf/aaai/GhoshB0D23} and post-hoc learning method~\cite{xi2024does}. There exist others concentrating on unique settings such as federated learning \citep{lu2023federated,DBLP:conf/icml/PlassierMRMP23}, multi-label problem \citep{cauchois2021knowing,fisch2022conformal,papadopoulos2014cross}, outlier detection \citep{bates2023testing,guan2022prediction}, and out-of-distribution detection~\cite{chen2023batch,novello2024out}. Most existing post-hoc methods calculate the non-conformity score based on unreliable probability values, leading to sub-optimal performance. In contrast to previous post-hoc methods, we show that probability value is not necessary for non-conformity scores and design an effective method to remove the probability value while retaining uncertainty information. 

\textbf{Validity of coverage rate.} There is an increasing amount of research focused on improving coverage rates\citep{6784618,lofstrom2015bias,ding2023class}, including efforts to maintain the marginal coverage rate by modifying the assumption of exchangeability to accommodate factors such as adversaries \citep{gendler2021adversarially,kang2024colep}, covariate shifts \citep{NEURIPS2019_8fb21ee7,deng2023happymap,xu2024robust},  label shifts \citep{podkopaev2021distribution,DBLP:conf/icml/PlassierMRMP23} and noisy labels \citep{einbinder2022conformal,sesia2023adaptive}. Moreover, conformal prediction algorithms pursue conditional coverage guarantee~\cite{vovk2012conditional}, including training-conditional validity and object-conditional validity. Specifically,
some methods acquire the training-conditional coverage guarantee which would ensure
that most draws of the training data set to result in valid marginal coverage
on future test points~\cite{bian2023training, pournaderi2024trainingconditional}.
There are also group-conditional conformal prediction methods that guarantee performance on all groups of individuals in the population~\cite{javanmard2022prediction,gibbs2023conformal,Melki_2023_ICCV}. 
Although it is impossible to obtain conditional coverage \cite{foygel2021limits}, THR \cite{sadinle2019least} shows that efficient class-conditional coverage is possible. For example, Clustered CP \cite{ding2023class} utilizes the taxonomy of the label space to improve the class-conditional coverage rate and $k$-Class-conditional CP calibrates the class-specific scores threshold depending on the top-$k$ error.
As discussed in Sec.\ref{sec:results}, SAPS broadly enhances the conditional coverage rate of prediction sets while maintaining small prediction sets.

\section{Conclusion}

In this paper, we present SAPS, a simple and effective conformal prediction score function that discards almost all the probability values except for the maximum softmax probability. By highlighting the information on label ranking, SAPS effectively mitigates the negative effect of tail probabilities, resulting in small and stable prediction sets. Additionally, we provide a key insight that the value of softmax probability might be redundant information for conformal prediction. Extensive experiments show that SAPS can improve the conditional coverage rate while maintaining a small prediction set. Our method is straightforward to implement with any classifier and can be easily adopted in various practical settings. We expect that out method can inspire future research to emphasize the label ranking information for conformal prediction.

\section{Limitations}
The limitation of our method is mainly on the hyperparameter representing the weight of label ranking that requires to be tuned with a validation dataset (same as RAPS). About future direction, we expect to enhance the conditional coverage of these conformal prediction algorithms on large-scale datasets, where all CP methods achieve worse conditional coverage rates than those of small-scale datasets. 

\section*{Acknowledgements}
This research is supported by the Shenzhen Fundamental Research Program (Grant No.
JCYJ20230807091809020). Yue Qiu is supported by the National Natural Science Foundation of China (Grant No. 12101407), the Chongqing Entrepreneurship and Innovation Program for Returned Overseas Scholars (Grant No. CX2023068), and the Fundamental Research Funds for the Central Universities (Grant No. 2022CDJJCLK-002, 2023CDJXY-042).
We gratefully acknowledge the support of the Center for Computational Science and Engineering at the Southern University of Science and Technology, as well as the Key Laboratory of Nonlinear Analysis and its Applications at Chongqing University, under the Ministry of Education, China, for our research. 

\section*{Impact Statement}
This paper presents work whose goal is to advance the field of Machine Learning. There are many potential societal consequences of our work, none of which we feel must be specifically highlighted here.

\clearpage


\bibliography{example_paper}

\begin{thebibliography}{82}
\providecommand{\natexlab}[1]{#1}
\providecommand{\url}[1]{\texttt{#1}}
\expandafter\ifx\csname urlstyle\endcsname\relax
  \providecommand{\doi}[1]{doi: #1}\else
  \providecommand{\doi}{doi: \begingroup \urlstyle{rm}\Url}\fi

\bibitem[Angelopoulos \& Bates(2021)Angelopoulos and Bates]{angelopoulos2021gentle}
Angelopoulos, A.~N. and Bates, S.
\newblock A gentle introduction to conformal prediction and distribution-free uncertainty quantification.
\newblock \emph{arXiv preprint arXiv:2107.07511}, 2021.

\bibitem[Angelopoulos et~al.(2021)Angelopoulos, Bates, Jordan, and Malik]{angelopoulos2020uncertainty}
Angelopoulos, A.~N., Bates, S., Jordan, M.~I., and Malik, J.
\newblock Uncertainty sets for image classifiers using conformal prediction.
\newblock In \emph{9th International Conference on Learning Representations, {ICLR} 2021, Virtual Event, Austria, May 3-7, 2021}. OpenReview.net, 2021.

\bibitem[Bai et~al.(2022)Bai, Mei, Wang, Zhou, and Xiong]{bai2022efficient}
Bai, Y., Mei, S., Wang, H., Zhou, Y., and Xiong, C.
\newblock Efficient and differentiable conformal prediction with general function classes.
\newblock \emph{arXiv preprint arXiv:2202.11091}, 2022.

\bibitem[Balasubramanian et~al.(2014)Balasubramanian, Ho, and Vovk]{balasubramanian2014conformal}
Balasubramanian, V., Ho, S.-S., and Vovk, V.
\newblock \emph{Conformal prediction for reliable machine learning: theory, adaptations and applications}.
\newblock Newnes, 2014.

\bibitem[Bang et~al.(2024)Bang, Dave, and Malikopoulos]{bang2024safe}
Bang, H., Dave, A., and Malikopoulos, A.~A.
\newblock Safe merging in mixed traffic with confidence.
\newblock \emph{arXiv preprint arXiv:2403.05742}, 2024.

\bibitem[Barber et~al.(2021)Barber, Cand{\`e}s, Ramdas, and Tibshirani]{barber2021predictive}
Barber, R.~F., Cand{\`e}s, E.~J., Ramdas, A., and Tibshirani, R.~J.
\newblock {Predictive inference with the jackknife+}.
\newblock \emph{The Annals of Statistics}, 49\penalty0 (1):\penalty0 486 -- 507, 2021.

\bibitem[Bates et~al.(2021)Bates, Angelopoulos, Lei, Malik, and Jordan]{bates2021distribution}
Bates, S., Angelopoulos, A., Lei, L., Malik, J., and Jordan, M.
\newblock Distribution-free, risk-controlling prediction sets.
\newblock \emph{Journal of the ACM (JACM)}, 68\penalty0 (6):\penalty0 1--34, 2021.

\bibitem[Bates et~al.(2023)Bates, Cand{\`e}s, Lei, Romano, and Sesia]{bates2023testing}
Bates, S., Cand{\`e}s, E., Lei, L., Romano, Y., and Sesia, M.
\newblock Testing for outliers with conformal p-values.
\newblock \emph{The Annals of Statistics}, 51\penalty0 (1):\penalty0 149--178, 2023.

\bibitem[Bellotti(2021)]{bellotti2021optimized}
Bellotti, A.
\newblock Optimized conformal classification using gradient descent approximation.
\newblock \emph{arXiv preprint arXiv:2105.11255}, 2021.

\bibitem[Bian \& Barber(2023)Bian and Barber]{bian2023training}
Bian, M. and Barber, R.~F.
\newblock Training-conditional coverage for distribution-free predictive inference.
\newblock \emph{Electronic Journal of Statistics}, pp.\  2044--2066, 2023.

\bibitem[Bojarski et~al.(2016)Bojarski, Del~Testa, Dworakowski, Firner, Flepp, Goyal, Jackel, Monfort, Muller, Zhang, et~al.]{bojarski2016end}
Bojarski, M., Del~Testa, D., Dworakowski, D., Firner, B., Flepp, B., Goyal, P., Jackel, L.~D., Monfort, M., Muller, U., Zhang, J., et~al.
\newblock End to end learning for self-driving cars.
\newblock \emph{arXiv preprint arXiv:1604.07316}, 2016.

\bibitem[Caruana et~al.(2015)Caruana, Lou, Gehrke, Koch, Sturm, and Elhadad]{caruana2015intelligible}
Caruana, R., Lou, Y., Gehrke, J., Koch, P., Sturm, M., and Elhadad, N.
\newblock Intelligible models for healthcare: Predicting pneumonia risk and hospital 30-day readmission.
\newblock In \emph{Proceedings of the 21th ACM SIGKDD International Conference on Knowledge Discovery and Data mining}, pp.\  1721--1730, 2015.

\bibitem[Cauchois et~al.(2021)Cauchois, Gupta, and Duchi]{cauchois2021knowing}
Cauchois, M., Gupta, S., and Duchi, J.~C.
\newblock Knowing what you know: valid and validated confidence sets in multiclass and multilabel prediction.
\newblock \emph{The Journal of Machine Learning Research}, 22\penalty0 (1):\penalty0 3681--3722, 2021.

\bibitem[Chen et~al.(2021)Chen, Huang, Lam, Qian, and Zhang]{chen2021learning}
Chen, H., Huang, Z., Lam, H., Qian, H., and Zhang, H.
\newblock Learning prediction intervals for regression: Generalization and calibration.
\newblock In \emph{International Conference on Artificial Intelligence and Statistics}, pp.\  820--828. PMLR, 2021.

\bibitem[Chen et~al.(2023)Chen, Li, and Yang]{chen2023batch}
Chen, X., Li, Y., and Yang, Y.
\newblock Batch-ensemble stochastic neural networks for out-of-distribution detection.
\newblock In \emph{ICASSP 2023-2023 IEEE International Conference on Acoustics, Speech and Signal Processing (ICASSP)}, pp.\  1--5. IEEE, 2023.

\bibitem[Clarkson(2023)]{clarkson2023distribution}
Clarkson, J.
\newblock Distribution free prediction sets for node classification.
\newblock In \emph{International Conference on Machine Learning}, pp.\  6268--6278. PMLR, 2023.

\bibitem[Colombo \& Vovk(2020)Colombo and Vovk]{colombo2020training}
Colombo, N. and Vovk, V.
\newblock Training conformal predictors.
\newblock In \emph{Conformal and Probabilistic Prediction and Applications}, pp.\  55--64. PMLR, 2020.

\bibitem[Cresswell et~al.(2024)Cresswell, Sui, Kumar, and Vouitsis]{cresswell2024conformal}
Cresswell, J.~C., Sui, Y., Kumar, B., and Vouitsis, N.
\newblock Conformal prediction sets improve human decision making.
\newblock \emph{arXiv preprint arXiv:2401.13744}, 2024.

\bibitem[Deng et~al.(2009)Deng, Dong, Socher, Li, Li, and Fei-Fei]{deng2009imagenet}
Deng, J., Dong, W., Socher, R., Li, L.-J., Li, K., and Fei-Fei, L.
\newblock Imagenet: A large-scale hierarchical image database.
\newblock In \emph{2009 IEEE Conference on Computer Vision and Pattern Recognition}, pp.\  248--255. Ieee, 2009.

\bibitem[Deng et~al.(2023)Deng, Dwork, and Zhang]{deng2023happymap}
Deng, Z., Dwork, C., and Zhang, L.
\newblock Happymap: A generalized multicalibration method.
\newblock In \emph{14th Innovations in Theoretical Computer Science Conference, ITCS 2023}, pp.\ ~41. Schloss Dagstuhl-Leibniz-Zentrum fur Informatik GmbH, Dagstuhl Publishing, 2023.

\bibitem[Ding et~al.(2023)Ding, Angelopoulos, Bates, Jordan, and Tibshirani]{ding2023class}
Ding, T., Angelopoulos, A.~N., Bates, S., Jordan, M.~I., and Tibshirani, R.~J.
\newblock Class-conditional conformal prediction with many classes.
\newblock \emph{arXiv preprint arXiv:2306.09335}, 2023.

\bibitem[Dosovitskiy et~al.(2020)Dosovitskiy, Beyer, Kolesnikov, Weissenborn, Zhai, Unterthiner, Dehghani, Minderer, Heigold, Gelly, et~al.]{dosovitskiy2020image}
Dosovitskiy, A., Beyer, L., Kolesnikov, A., Weissenborn, D., Zhai, X., Unterthiner, T., Dehghani, M., Minderer, M., Heigold, G., Gelly, S., et~al.
\newblock An image is worth 16x16 words: Transformers for image recognition at scale.
\newblock \emph{arXiv preprint arXiv:2010.11929}, 2020.

\bibitem[Einbinder et~al.(2022)Einbinder, Romano, Sesia, and Zhou]{einbinder2022training}
Einbinder, B.-S., Romano, Y., Sesia, M., and Zhou, Y.
\newblock Training uncertainty-aware classifiers with conformalized deep learning.
\newblock \emph{Advances in Neural Information Processing Systems}, 35:\penalty0 22380--22395, 2022.

\bibitem[Feldman et~al.(2023)Feldman, Einbinder, Bates, Angelopoulos, Gendler, and Romano]{einbinder2022conformal}
Feldman, S., Einbinder, B.-S., Bates, S., Angelopoulos, A.~N., Gendler, A., and Romano, Y.
\newblock Conformal prediction is robust to dispersive label noise.
\newblock In \emph{Conformal and Probabilistic Prediction with Applications}, pp.\  624--626. PMLR, 2023.

\bibitem[Fisch et~al.(2021)Fisch, Schuster, Jaakkola, and Barzilay]{fisch2021few}
Fisch, A., Schuster, T., Jaakkola, T., and Barzilay, R.
\newblock Few-shot conformal prediction with auxiliary tasks.
\newblock In \emph{International Conference on Machine Learning}, pp.\  3329--3339. PMLR, 2021.

\bibitem[Fisch et~al.(2022)Fisch, Schuster, Jaakkola, and Barzilay]{fisch2022conformal}
Fisch, A., Schuster, T., Jaakkola, T., and Barzilay, R.
\newblock Conformal prediction sets with limited false positives.
\newblock In \emph{International Conference on Machine Learning}, pp.\  6514--6532. PMLR, 2022.

\bibitem[Foygel~Barber et~al.(2021)Foygel~Barber, Candes, Ramdas, and Tibshirani]{foygel2021limits}
Foygel~Barber, R., Candes, E.~J., Ramdas, A., and Tibshirani, R.~J.
\newblock The limits of distribution-free conditional predictive inference.
\newblock \emph{Information and Inference: A Journal of the IMA}, 10\penalty0 (2):\penalty0 455--482, 2021.

\bibitem[Gal \& Ghahramani(2016)Gal and Ghahramani]{gal2016dropout}
Gal, Y. and Ghahramani, Z.
\newblock Dropout as a bayesian approximation: Representing model uncertainty in deep learning.
\newblock In \emph{International Conference on Machine Learning}, pp.\  1050--1059. PMLR, 2016.

\bibitem[Gendler et~al.(2021)Gendler, Weng, Daniel, and Romano]{gendler2021adversarially}
Gendler, A., Weng, T.-W., Daniel, L., and Romano, Y.
\newblock Adversarially robust conformal prediction.
\newblock In \emph{International Conference on Learning Representations}, 2021.

\bibitem[Ghosh et~al.(2023)Ghosh, Belkhouja, Yan, and Doppa]{DBLP:conf/aaai/GhoshB0D23}
Ghosh, S., Belkhouja, T., Yan, Y., and Doppa, J.~R.
\newblock Improving uncertainty quantification of deep classifiers via neighborhood conformal prediction: Novel algorithm and theoretical analysis.
\newblock In \emph{Proceedings of the AAAI Conference on Artificial Intelligence}, pp.\  7722--7730. {AAAI} Press, 2023.

\bibitem[Gibbs et~al.(2023)Gibbs, Cherian, and Cand{\`e}s]{gibbs2023conformal}
Gibbs, I., Cherian, J.~J., and Cand{\`e}s, E.~J.
\newblock Conformal prediction with conditional guarantees.
\newblock \emph{arXiv preprint arXiv:2305.12616}, 2023.

\bibitem[Guan \& Tibshirani(2022)Guan and Tibshirani]{guan2022prediction}
Guan, L. and Tibshirani, R.
\newblock Prediction and outlier detection in classification problems.
\newblock \emph{Journal of the Royal Statistical Society Series B: Statistical Methodology}, 84\penalty0 (2):\penalty0 524--546, 2022.

\bibitem[Guo et~al.(2017)Guo, Pleiss, Sun, and Weinberger]{guo2017calibration}
Guo, C., Pleiss, G., Sun, Y., and Weinberger, K.~Q.
\newblock On calibration of modern neural networks.
\newblock In \emph{International Conference on Machine Learning}, pp.\  1321--1330. PMLR, 2017.

\bibitem[Hendrycks \& Gimpel(2016)Hendrycks and Gimpel]{hendrycks2016baseline}
Hendrycks, D. and Gimpel, K.
\newblock A baseline for detecting misclassified and out-of-distribution examples in neural networks.
\newblock In \emph{International Conference on Learning Representations}, 2016.

\bibitem[Horwitz \& Hoshen(2022)Horwitz and Hoshen]{horwitz2022conffusion}
Horwitz, E. and Hoshen, Y.
\newblock Conffusion: Confidence intervals for diffusion models.
\newblock \emph{arXiv preprint arXiv:2211.09795}, 2022.

\bibitem[Javanmard et~al.(2022)Javanmard, Shao, and Bien]{javanmard2022prediction}
Javanmard, A., Shao, S., and Bien, J.
\newblock Prediction sets for high-dimensional mixture of experts models.
\newblock \emph{arXiv preprint arXiv:2210.16710}, 2022.

\bibitem[Kang et~al.(2024)Kang, G{\"u}rel, Li, and Li]{kang2024colep}
Kang, M., G{\"u}rel, N.~M., Li, L., and Li, B.
\newblock Colep: Certifiably robust learning-reasoning conformal prediction via probabilistic circuits.
\newblock \emph{arXiv preprint arXiv:2403.11348}, 2024.

\bibitem[Krizhevsky et~al.(2009)Krizhevsky, Hinton, et~al.]{krizhevsky2009learning}
Krizhevsky, A., Hinton, G., et~al.
\newblock Learning multiple layers of features from tiny images.
\newblock 2009.

\bibitem[Kumar et~al.(2023)Kumar, Lu, Gupta, Palepu, Bellamy, Raskar, and Beam]{kumar2023conformal}
Kumar, B., Lu, C., Gupta, G., Palepu, A., Bellamy, D., Raskar, R., and Beam, A.
\newblock Conformal prediction with large language models for multi-choice question answering.
\newblock \emph{arXiv preprint arXiv:2305.18404}, 2023.

\bibitem[Lei \& Wasserman(2014)Lei and Wasserman]{lei2014distribution}
Lei, J. and Wasserman, L.
\newblock Distribution-free prediction bands for non-parametric regression.
\newblock \emph{Journal of the Royal Statistical Society Series B: Statistical Methodology}, 76\penalty0 (1):\penalty0 71--96, 2014.

\bibitem[Lei et~al.(2015)Lei, Rinaldo, and Wasserman]{lei2015conformal}
Lei, J., Rinaldo, A., and Wasserman, L.
\newblock A conformal prediction approach to explore functional data.
\newblock \emph{Annals of Mathematics and Artificial Intelligence}, 74:\penalty0 29--43, 2015.

\bibitem[Liang et~al.(2023)Liang, Zhou, and Sesia]{DBLP:conf/icml/LiangZS23}
Liang, Z., Zhou, Y., and Sesia, M.
\newblock Conformal inference is (almost) free for neural networks trained with early stopping.
\newblock In \emph{International Conference on Machine Learning}, volume 202 of \emph{Proceedings of Machine Learning Research}, pp.\  20810--20851. {PMLR}, 2023.

\bibitem[L{\"o}fstr{\"o}m et~al.(2015)L{\"o}fstr{\"o}m, Bostr{\"o}m, Linusson, and Johansson]{lofstrom2015bias}
L{\"o}fstr{\"o}m, T., Bostr{\"o}m, H., Linusson, H., and Johansson, U.
\newblock Bias reduction through conditional conformal prediction.
\newblock \emph{Intelligent Data Analysis}, 19\penalty0 (6):\penalty0 1355--1375, 2015.

\bibitem[Lu et~al.(2023)Lu, Yu, Karimireddy, Jordan, and Raskar]{lu2023federated}
Lu, C., Yu, Y., Karimireddy, S.~P., Jordan, M., and Raskar, R.
\newblock Federated conformal predictors for distributed uncertainty quantification.
\newblock In \emph{International Conference on Machine Learning}, pp.\  22942--22964. PMLR, 2023.

\bibitem[Melki et~al.(2023)Melki, Bombrun, Diallo, Dias, and Da~Costa]{Melki_2023_ICCV}
Melki, P., Bombrun, L., Diallo, B., Dias, J., and Da~Costa, J.-P.
\newblock Group-conditional conformal prediction via quantile regression calibration for crop and weed classification.
\newblock In \emph{Proceedings of the IEEE/CVF International Conference on Computer Vision (ICCV) Workshops}, pp.\  614--623, October 2023.

\bibitem[Moya et~al.(2024)Moya, Mollaali, Zhang, Lu, and Lin]{moya2024conformalized}
Moya, C., Mollaali, A., Zhang, Z., Lu, L., and Lin, G.
\newblock Conformalized-deeponet: A distribution-free framework for uncertainty quantification in deep operator networks.
\newblock \emph{arXiv preprint arXiv:2402.15406}, 2024.

\bibitem[Novello et~al.(2024)Novello, Dalmau, and Andeol]{novello2024out}
Novello, P., Dalmau, J., and Andeol, L.
\newblock Out-of-distribution detection should use conformal prediction (and vice-versa?).
\newblock \emph{arXiv preprint arXiv:2403.11532}, 2024.

\bibitem[Papadopoulos(2014)]{papadopoulos2014cross}
Papadopoulos, H.
\newblock A cross-conformal predictor for multi-label classification.
\newblock In \emph{Artificial Intelligence Applications and Innovations: AIAI 2014 Workshops: CoPA, MHDW, IIVC, and MT4BD, Rhodes, Greece, September 19-21, 2014. Proceedings 10}, pp.\  241--250. Springer, 2014.

\bibitem[Papadopoulos et~al.(2002{\natexlab{a}})Papadopoulos, Proedrou, Vovk, and Gammerman]{10.1007/3-540-36755-1_29}
Papadopoulos, H., Proedrou, K., Vovk, V., and Gammerman, A.
\newblock Inductive confidence machines for regression.
\newblock In \emph{Machine Learning: ECML 2002}, pp.\  345--356. Springer Berlin Heidelberg, 2002{\natexlab{a}}.

\bibitem[Papadopoulos et~al.(2002{\natexlab{b}})Papadopoulos, Vovk, and Gammerman]{Papadopoulos2002QualifiedPF}
Papadopoulos, H., Vovk, V., and Gammerman, A.
\newblock Qualified prediction for large data sets in the case of pattern recognition.
\newblock In \emph{International Conference on Machine Learning and Applications}, 2002{\natexlab{b}}.

\bibitem[Plassier et~al.(2023)Plassier, Makni, Rubashevskii, Moulines, and Panov]{DBLP:conf/icml/PlassierMRMP23}
Plassier, V., Makni, M., Rubashevskii, A., Moulines, E., and Panov, M.
\newblock Conformal prediction for federated uncertainty quantification under label shift.
\newblock In \emph{International Conference on Machine Learning}, volume 202 of \emph{Proceedings of Machine Learning Research}, pp.\  27907--27947. {PMLR}, 2023.

\bibitem[Podina et~al.(2024)Podina, Rad, and Kohandel]{podina2024conformalized}
Podina, L., Rad, M.~T., and Kohandel, M.
\newblock Conformalized physics-informed neural networks.
\newblock \emph{arXiv preprint arXiv:2405.08111}, 2024.

\bibitem[Podkopaev \& Ramdas(2021)Podkopaev and Ramdas]{podkopaev2021distribution}
Podkopaev, A. and Ramdas, A.
\newblock Distribution-free uncertainty quantification for classification under label shift.
\newblock In \emph{Uncertainty in Artificial Intelligence}, pp.\  844--853. PMLR, 2021.

\bibitem[Pournaderi \& Xiang(2024)Pournaderi and Xiang]{pournaderi2024trainingconditional}
Pournaderi, M. and Xiang, Y.
\newblock Training-conditional coverage bounds under covariate shift, 2024.

\bibitem[Quach et~al.(2023)Quach, Fisch, Schuster, Yala, Sohn, Jaakkola, and Barzilay]{quach2023conformal}
Quach, V., Fisch, A., Schuster, T., Yala, A., Sohn, J.~H., Jaakkola, T.~S., and Barzilay, R.
\newblock Conformal language modeling.
\newblock \emph{arXiv preprint arXiv:2306.10193}, 2023.

\bibitem[Radford et~al.(2021)Radford, Kim, Hallacy, Ramesh, Goh, Agarwal, Sastry, Askell, Mishkin, Clark, et~al.]{radford2021learning}
Radford, A., Kim, J.~W., Hallacy, C., Ramesh, A., Goh, G., Agarwal, S., Sastry, G., Askell, A., Mishkin, P., Clark, J., et~al.
\newblock Learning transferable visual models from natural language supervision.
\newblock In \emph{International Conference on Machine Learning}, pp.\  8748--8763. PMLR, 2021.

\bibitem[Recht et~al.(2019)Recht, Roelofs, Schmidt, and Shankar]{recht2019imagenet}
Recht, B., Roelofs, R., Schmidt, L., and Shankar, V.
\newblock Do imagenet classifiers generalize to imagenet?
\newblock In \emph{International Conference on Machine Learning}, pp.\  5389--5400. PMLR, 2019.

\bibitem[Ren et~al.(2023)Ren, Dixit, Bodrova, Singh, Tu, Brown, Xu, Takayama, Xia, Varley, et~al.]{ren2023robots}
Ren, A.~Z., Dixit, A., Bodrova, A., Singh, S., Tu, S., Brown, N., Xu, P., Takayama, L., Xia, F., Varley, J., et~al.
\newblock Robots that ask for help: Uncertainty alignment for large language model planners.
\newblock In \emph{Conference on Robot Learning}, pp.\  661--682. PMLR, 2023.

\bibitem[Romano et~al.(2019)Romano, Patterson, and Candes]{romano2019conformalized}
Romano, Y., Patterson, E., and Candes, E.
\newblock Conformalized quantile regression.
\newblock \emph{Advances in Neural Information Processing Systems}, 32, 2019.

\bibitem[Romano et~al.(2020)Romano, Sesia, and Candes]{romano2020classification}
Romano, Y., Sesia, M., and Candes, E.
\newblock Classification with valid and adaptive coverage.
\newblock \emph{Advances in Neural Information Processing Systems}, 33:\penalty0 3581--3591, 2020.

\bibitem[Sadinle et~al.(2019)Sadinle, Lei, and Wasserman]{sadinle2019least}
Sadinle, M., Lei, J., and Wasserman, L.
\newblock Least ambiguous set-valued classifiers with bounded error levels.
\newblock \emph{Journal of the American Statistical Association}, 114\penalty0 (525):\penalty0 223--234, 2019.

\bibitem[Sesia et~al.(2023)Sesia, Wang, and Tong]{sesia2023adaptive}
Sesia, M., Wang, Y.~R., and Tong, X.
\newblock Adaptive conformal classification with noisy labels.
\newblock In \emph{2023 IMS International Conference on Statistics and Data Science (ICSDS)}, pp.\  221, 2023.

\bibitem[Shafer \& Vovk(2008)Shafer and Vovk]{shafer2008tutorial}
Shafer, G. and Vovk, V.
\newblock A tutorial on conformal prediction.
\newblock \emph{Journal of Machine Learning Research}, 9\penalty0 (3), 2008.

\bibitem[Shi et~al.(2013)Shi, Ong, and Leckie]{6784618}
Shi, F., Ong, C.~S., and Leckie, C.
\newblock Applications of class-conditional conformal predictor in multi-class classification.
\newblock In \emph{2013 12th International Conference on Machine Learning and Applications}, volume~1, pp.\  235--239, 2013.
\newblock \doi{10.1109/ICMLA.2013.48}.

\bibitem[Smith(2013)]{smith2013uncertainty}
Smith, R.~C.
\newblock \emph{Uncertainty quantification: theory, implementation, and applications}, volume~12.
\newblock Siam, 2013.

\bibitem[Song et~al.(2024)Song, Huang, Jiang, Zhang, Li, and Wang]{song2024similarity}
Song, J., Huang, J., Jiang, W., Zhang, B., Li, S., and Wang, C.
\newblock Similarity-navigated conformal prediction for graph neural networks.
\newblock \emph{arXiv preprint arXiv:2405.14303}, 2024.

\bibitem[Straitouri et~al.(2023)Straitouri, Wang, Okati, and Rodriguez]{straitouri2022improving}
Straitouri, E., Wang, L., Okati, N., and Rodriguez, M.~G.
\newblock Improving expert predictions with conformal prediction.
\newblock In \emph{Proceedings of the 40th International Conference on Machine Learning}, ICML'23. JMLR.org, 2023.

\bibitem[Stutz et~al.(2022)Stutz, Dvijotham, Cemgil, and Doucet]{DBLP:conf/iclr/StutzDCD22}
Stutz, D., Dvijotham, K., Cemgil, A.~T., and Doucet, A.
\newblock Learning optimal conformal classifiers.
\newblock In \emph{The Tenth International Conference on Learning Representations, {ICLR} 2022, Virtual Event, April 25-29, 2022}. OpenReview.net, 2022.

\bibitem[Su et~al.(2024)Su, Luo, Wang, and Cheng]{su2024api}
Su, J., Luo, J., Wang, H., and Cheng, L.
\newblock Api is enough: Conformal prediction for large language models without logit-access.
\newblock \emph{arXiv preprint arXiv:2403.01216}, 2024.

\bibitem[Teneggi et~al.(2023)Teneggi, Tivnan, Stayman, and Sulam]{teneggi2023trust}
Teneggi, J., Tivnan, M., Stayman, W., and Sulam, J.
\newblock How to trust your diffusion model: A convex optimization approach to conformal risk control.
\newblock In \emph{International Conference on Machine Learning}, pp.\  33940--33960. PMLR, 2023.

\bibitem[Teng et~al.(2023)Teng, Wen, Zhang, Bengio, Gao, and Yuan]{DBLP:conf/iclr/TengWZB0Y23}
Teng, J., Wen, C., Zhang, D., Bengio, Y., Gao, Y., and Yuan, Y.
\newblock Predictive inference with feature conformal prediction.
\newblock In \emph{The Eleventh International Conference on Learning Representations, {ICLR} 2023, Kigali, Rwanda, May 1-5, 2023}, 2023.

\bibitem[Tibshirani et~al.(2019)Tibshirani, Foygel~Barber, Candes, and Ramdas]{NEURIPS2019_8fb21ee7}
Tibshirani, R.~J., Foygel~Barber, R., Candes, E., and Ramdas, A.
\newblock Conformal prediction under covariate shift.
\newblock In \emph{Advances in Neural Information Processing Systems}, volume~32. Curran Associates, Inc., 2019.

\bibitem[Touvron et~al.(2021)Touvron, Cord, Douze, Massa, Sablayrolles, and J{\'e}gou]{touvron2021training}
Touvron, H., Cord, M., Douze, M., Massa, F., Sablayrolles, A., and J{\'e}gou, H.
\newblock Training data-efficient image transformers \& distillation through attention.
\newblock In \emph{International Conference on Machine Learning}, pp.\  10347--10357. PMLR, 2021.

\bibitem[Vovk(2002)]{1181895}
Vovk, V.
\newblock On-line confidence machines are well-calibrated.
\newblock In \emph{The 43rd Annual IEEE Symposium on Foundations of Computer Science, 2002. Proceedings.}, pp.\  187--196, 2002.

\bibitem[Vovk(2012)]{vovk2012conditional}
Vovk, V.
\newblock Conditional validity of inductive conformal predictors.
\newblock In \emph{Asian Conference on Machine Learning}, pp.\  475--490. PMLR, 2012.

\bibitem[Vovk(2015)]{vovk2015cross}
Vovk, V.
\newblock Cross-conformal predictors.
\newblock \emph{Annals of Mathematics and Artificial Intelligence}, 74:\penalty0 9--28, 2015.

\bibitem[Vovk et~al.(2005)Vovk, Gammerman, and Shafer]{vovk2005algorithmic}
Vovk, V., Gammerman, A., and Shafer, G.
\newblock \emph{Algorithmic learning in a random world}, volume~29.
\newblock Springer, 2005.

\bibitem[Wang et~al.(2023)Wang, Tong, Tan, Vorobeychik, and Kantaros]{wang2023conformal}
Wang, J., Tong, J., Tan, K., Vorobeychik, Y., and Kantaros, Y.
\newblock Conformal temporal logic planning using large language models: Knowing when to do what and when to ask for help.
\newblock \emph{arXiv preprint arXiv:2309.10092}, 2023.

\bibitem[Wijegunawardana et~al.(2020)Wijegunawardana, Gera, and Soundarajan]{wijegunawardana2020node}
Wijegunawardana, P., Gera, R., and Soundarajan, S.
\newblock Node classification with bounded error rates.
\newblock In \emph{Complex Networks XI: Proceedings of the 11th Conference on Complex Networks CompleNet 2020}, pp.\  26--38. Springer, 2020.

\bibitem[Xi et~al.(2024)Xi, Huang, Feng, and Wei]{xi2024does}
Xi, H., Huang, J., Feng, L., and Wei, H.
\newblock Does confidence calibration help conformal prediction?
\newblock \emph{arXiv preprint arXiv:2402.04344}, 2024.

\bibitem[Xu et~al.(2024)Xu, Sun, Chen, Venkitasubramaniam, and Xie]{xu2024robust}
Xu, R., Sun, Y., Chen, C., Venkitasubramaniam, P., and Xie, S.
\newblock Robust conformal prediction under distribution shift via physics-informed structural causal model.
\newblock \emph{arXiv preprint arXiv:2403.15025}, 2024.

\bibitem[Yang \& Kuchibhotla(2021)Yang and Kuchibhotla]{yang2021finite}
Yang, Y. and Kuchibhotla, A.~K.
\newblock Finite-sample efficient conformal prediction.
\newblock \emph{arXiv preprint arXiv:2104.13871}, 2021.

\end{thebibliography}
\bibliographystyle{icml2024}

\newpage
\appendix
\onecolumn

\section{Proof of Proposition~\ref{proposition:aps_constant}}
\label{appendix:aps_constant}

\begin{proof}
    Let $\hat{\pi}$ be a classifier.
    $(\boldsymbol{x}_1,y_1)$ and $(\boldsymbol{x}_2,y_2)$ represent two instances sampled from the joint data distribution $\mathcal{P}_{\mathcal{XY}}$, and $u_1,u_2$ are sampled from $U[0,1]$.
    It is easy to get that for any different positive numbers $\gamma_1$ and $\gamma_2$, $\gamma_1 \cdot [o(y_1,\hat{\pi}(\boldsymbol{x}_1))-1+ u_1] \leq \gamma_1 \cdot [o(y_2,\hat{\pi}(\boldsymbol{x}_2))-1+ u_2]$ if and only if $\gamma_2 \cdot [o(y_1,\hat{\pi}(\boldsymbol{x}_1))-1+ u_1] \leq \gamma_2 \cdot [o(y_2,\hat{\pi}(\boldsymbol{x}_2))-1+ u_2]$. According to Lemma~\ref{definition:equivalent_scores}, we can get that $S_{cons}(\boldsymbol{x},y;\hat{\pi},\gamma_1)$ and $S_{cons}(\boldsymbol{x},y;\hat{\pi},\gamma_2)$ could provide the same prediction sets.
\end{proof}

\section{Proof of Theorem~\ref{theorem:constantaps}}
\label{proof:theorem:constantaps}
\begin{proof}
    For simplicity, let $a_r=A_{r}-A_{r-1}$  for $r\geq 1$ and $A_0 = 0$. $a_r$ represents the rate of examples in the calibration set whose ground-truth labels' rank is $r$.
    The non-conformity scores of examples whose rank of ground-truth labels are $r$ can be computed by 
    $$ S_{cons}(\boldsymbol{x},y,u;\hat{\pi}) =  r -u ,$$
    where $u \sim U[0,1]$.
    Thus, we can observe that the scores of these examples are uniformly distributed on $[r-1,r]$.  Scores of examples whose rank of ground-truth labels are less than or equal to $r$  are no more than $r$. In other words, the rate of examples in the calibration set with scores lower than $r$ is $A_{r}$. 
      Given that $k$ satisfies $A_{k}\geq 1-\alpha >A_{k-1}$, the $1-\alpha$ quantile of scores for the calibration set, i.e., the calibrated threshold $\tau$, is located in the interval $[k-1,k]$. Since the scores in the interval $[k-1,k]$ are uniformly distributed and the rate of examples whose scores in $[k-1,k]$ is $a_k$, $\tau$ is equivalent to the $\frac{1-\alpha- A_{k-1}}{a_{k}}$ quantile of $[k-1,k]$. Formally, we can get the value of $\tau$ by
    $$\tau = (k-1) +\frac{1-\alpha- A_{k-1}}{a_{k}}.$$
    Based on the definition of prediction set in Equation~\ref{eq:cp_set}, the prediction set of  a test instance $\boldsymbol{x}$ is equal to 
    \begin{equation}
        \mathcal{C}_{1-\alpha}(\boldsymbol{x},u) =\{y\in\mathcal{Y}: o(y,\hat{\pi}(\boldsymbol{x}))-u\leq \tau\}.
    \end{equation}
    Next, we analyze the prediction set $ \mathcal{C}_{1-\alpha}(\boldsymbol{x},u)$.
    When $y\in\mathcal{Y}$ satisfies the inequation $o(y,\hat{\pi}(\boldsymbol{x}))\leq k-1$, $S_{cons}(\boldsymbol{x},y,u;\hat{\pi})$ must be smaller than $ \tau$. Thus, the size of sets is at least $k-1$. But for  $y$ satisfying $o(y,\hat{\pi}(\boldsymbol{x}))= k$, we can have
    \begin{align*}
        o(y,\hat{\pi}(\boldsymbol{x}))-u  &\leq \tau \\
        k -u  &\leq (k-1) +\frac{1-\alpha- A_{k-1}}{a_{k}} \\
        1-\frac{1-\alpha- A_{k-1}}{a_{k}} &\leq u \\
    \end{align*}
    Thus, there exists a probability of $\frac{1-\alpha- A_{k-1}}{a_{k}}$ such that $S(\boldsymbol{x},y,u;\hat{\pi})  \leq \tau$.
    Finally, the expected value of the set size for the test example $\boldsymbol{x}$ is 
    \begin{equation}
        \mathbb{E}_{u\sim U}[|\mathcal{C}_{1-\alpha}(\boldsymbol{x},u)|] = k-1 +\frac{1-\alpha- A_{k-1}}{a_{k}} = k-1+\frac{1-\alpha - A_{k-1}}{A_{k}-A_{k-1}}
    \end{equation}
\end{proof}

\section{Detailed results for coverage rate and set size}
\label{appendix:set_size}

In this section, we report the detailed results of coverage rate and set size on different datasets in Table~\ref{table:size_imagenet},\ref{table:size_cifar100}, and \ref{table:size_cifar10}.  The median-of-means for each result is reported over ten different trials.  The average results of multiple models have been reported in the main paper.

\begin{table}[H] 
\centering 
\caption{Results on ImageNet. The median-of-means for each column is reported over 10 different trials.  \textbf{Bold} numbers indicate optimal performance.} 
\label{table:size_imagenet} 
\resizebox{0.95\textwidth}{!}{
\setlength{\tabcolsep}{2mm}{ 
\begin{tabular}{lcccccccccccc } 
\toprule 
 & \multicolumn{6}{c}{$\alpha=0.1$}  & \multicolumn{6}{c}{$\alpha=0.05$} \\ 
\cmidrule(r){2-7}  \cmidrule(r){8-13}  
 & \multicolumn{3}{c}{Coverage}  & \multicolumn{3}{c}{Size $\downarrow$} & \multicolumn{3}{c}{Coverage}  & \multicolumn{3}{c}{Size $\downarrow$}\\ 
\cmidrule(r){2-4}  \cmidrule(r){5-7} \cmidrule(r){8-10}  \cmidrule(r){11-13}  
Datasets   & APS & RAPS &SAPS & APS & RAPS &SAPS & APS & RAPS &SAPS & APS & RAPS &SAPS\\ 
\midrule 
 ResNeXt101 & 0.899 & 0.902 & 0.901 & 19.49 & 2.01 & \textbf{1.82}  &0.950 & 0.951 & 0.950 & 46.58 & 4.24 & \textbf{3.83}  \\ 
 ResNet152 & 0.900 & 0.900 & 0.900 & 10.51 & 2.10 & \textbf{1.92}  &0.950 & 0.950 & 0.950 & 22.65 & 4.39 & \textbf{4.07}  \\ 
 ResNet101 & 0.898 & 0.900 & 0.900 & 10.83 & 2.24 & \textbf{2.07}  &0.948 & 0.949 & 0.950 & 23.20 & 4.78 & \textbf{4.34}  \\ 
 ResNet50 & 0.899 & 0.900 & 0.900 & 12.29 & 2.51 & \textbf{2.31}  &0.948 & 0.950 & 0.950 & 25.99 & 5.57 & \textbf{5.25}  \\ 
 ResNet18 & 0.899 & 0.900 & 0.900 & 16.10 & 4.43 & \textbf{4.00}  &0.949 & 0.950 & 0.950 & 32.89 & 11.75 & \textbf{10.47}  \\ 
 DenseNet161 & 0.900 & 0.900 & 0.900 & 12.03 & 2.27 & \textbf{2.08}  &0.949 & 0.950 & 0.951 & 28.06 & 5.11 & \textbf{4.61}  \\ 
 VGG16 & 0.897 & 0.901 & 0.900 & 14.00 & 3.59 & \textbf{3.25}  &0.948 & 0.950 & 0.949 & 27.55 & 8.80 & \textbf{7.84}  \\ 
 Inception & 0.900 & 0.902 & 0.902 & 87.93 & 5.32 & \textbf{4.58}  &0.949 & 0.951 & 0.950 & 167.98 & 18.71 & \textbf{14.43}  \\ 
 ShuffleNet & 0.900 & 0.899 & 0.900 & 31.77 & 5.04 & \textbf{4.54}  &0.949 & 0.950 & 0.950 & 69.39 & 16.13 & \textbf{14.05}  \\ 
 ViT & 0.900 & 0.898 & 0.900 & 10.55 & 1.70 & \textbf{1.61}  &0.950 & 0.949 & 0.950 & 31.75 & 3.91 & \textbf{3.21}  \\ 
 DeiT & 0.901 & 0.900 & 0.900 & 8.51 & 1.48 & \textbf{1.41}  &0.950 & 0.949 & 0.949 & 24.88 & 2.69 & \textbf{2.49}  \\ 
 CLIP & 0.899 & 0.900 & 0.900 & 17.45 & 6.81 & \textbf{6.23}  &0.951 & 0.949 & 0.949 & 35.09 & 16.79 & \textbf{16.07}  \\ 
\midrule 
 average & 0.899 & 0.900 & 0.900 & 20.95 & 3.29 & \textbf{2.98}  &0.949 & 0.950 & 0.950 & 44.67 & 8.57 & \textbf{7.55}  \\ 
\bottomrule 
\end{tabular}}} 
\end{table}

\begin{table}[H] 
\centering 
\caption{Results on CIFAR-100. The median-of-means for each column is reported over 10 different trials.  \textbf{Bold} numbers indicate optimal performance. } 
\label{table:size_cifar100} 
\resizebox{0.95\textwidth}{!}{
\setlength{\tabcolsep}{2mm}{ 
\begin{tabular}{lcccccccccccc } 
\toprule 
 & \multicolumn{6}{c}{$\alpha=0.1$}  & \multicolumn{6}{c}{$\alpha=0.05$} \\ 
\cmidrule(r){2-7}  \cmidrule(r){8-13}  
 & \multicolumn{3}{c}{Coverage}  & \multicolumn{3}{c}{Size $\downarrow$} & \multicolumn{3}{c}{Coverage}  & \multicolumn{3}{c}{Size $\downarrow$}\\ 
\cmidrule(r){2-4}  \cmidrule(r){5-7} \cmidrule(r){8-10}  \cmidrule(r){11-13}  
Datasets   & APS & RAPS &SAPS & APS & RAPS &SAPS & APS & RAPS &SAPS & APS & RAPS &SAPS\\ 
\midrule 
 ResNet18 & 0.898 & 0.901 & 0.898 & 10.03 & 2.72 & \textbf{2.41}  &0.949 & 0.950 & 0.951 & 16.76 & 5.88 & \textbf{4.96}  \\ 
 ResNet50 & 0.896 & 0.902 & 0.899 & 6.51 & 2.16 & \textbf{2.04}  &0.946 & 0.948 & 0.946 & 12.49 & 4.37 & \textbf{3.69}  \\ 
 ResNet101 & 0.899 & 0.901 & 0.898 & 6.52 & 2.10 & \textbf{1.99}  &0.951 & 0.947 & 0.948 & 12.26 & 4.49 & \textbf{3.85}  \\ 
 DenseNet161 & 0.898 & 0.898 & 0.897 & 8.07 & 2.02 & \textbf{1.77}  &0.948 & 0.949 & 0.950 & 14.35 & 3.61 & \textbf{3.33}  \\ 
 VGG16 & 0.900 & 0.896 & 0.895 & 5.80 & 4.58 & \textbf{3.64}  &0.951 & 0.949 & 0.948 & 11.83 & 11.32 & \textbf{9.27}  \\ 
 Inception & 0.902 & 0.902 & 0.902 & 12.01 & 2.01 & \textbf{2.01}  &0.952 & 0.953 & 0.953 & 18.24 & 5.21 & \textbf{4.26}  \\ 
 ViT & 0.897 & 0.901 & 0.900 & 4.29 & 2.14 & \textbf{1.91}  &0.949 & 0.948 & 0.949 & 7.92 & 3.98 & \textbf{3.57}  \\ 
 CLIP & 0.899 & 0.900 & 0.900 & 9.84 & 6.18 & \textbf{5.58}  &0.952 & 0.948 & 0.950 & 16.04 & 12.50 & \textbf{11.27}  \\ 
\midrule 
 average & 0.899 & 0.900 & 0.899 & 7.88 & 2.99 & \textbf{2.67}  &0.950 & 0.949 & 0.949 & 13.74 & 6.42 & \textbf{5.53}  \\ 
\bottomrule 
\end{tabular}}} 
\end{table}

\begin{table}[H] 
\centering 
\caption{Results on CIFAR-10. The median-of-means for each column is reported over 10 different trials.  \textbf{Bold} numbers indicate optimal performance. } 
\label{table:size_cifar10} 
\resizebox{0.95\textwidth}{!}{
\setlength{\tabcolsep}{2mm}{ 
\begin{tabular}{lcccccccccccc } 
\toprule 
 & \multicolumn{6}{c}{$\alpha=0.1$}  & \multicolumn{6}{c}{$\alpha=0.05$} \\ 
\cmidrule(r){2-7}  \cmidrule(r){8-13}  
 & \multicolumn{3}{c}{Coverage}  & \multicolumn{3}{c}{Size $\downarrow$} & \multicolumn{3}{c}{Coverage}  & \multicolumn{3}{c}{Size $\downarrow$}\\ 
\cmidrule(r){2-4}  \cmidrule(r){5-7} \cmidrule(r){8-10}  \cmidrule(r){11-13}  
Datasets   & APS & RAPS &SAPS & APS & RAPS &SAPS & APS & RAPS &SAPS & APS & RAPS &SAPS\\ 
\midrule 
 ResNet18 & 0.896 & 0.896 & 0.898 & 2.42 & 2.29 & \textbf{2.18}  &0.948 & 0.947 & 0.949 & 3.12 & 3.12 & \textbf{3.10}  \\ 
 ResNet50 & 0.897 & 0.900 & 0.895 & 2.08 & 1.95 & \textbf{1.77}  &0.949 & 0.950 & 0.950 & 2.69 & 2.62 & \textbf{2.53}  \\ 
 ResNet101 & 0.899 & 0.901 & 0.898 & 2.15 & 2.00 & \textbf{1.86}  &0.949 & 0.948 & 0.949 & 2.78 & 2.68 & \textbf{2.59}  \\ 
 DenseNet161 & 0.898 & 0.899 & 0.898 & 2.06 & 1.90 & \textbf{1.71}  &0.949 & 0.953 & 0.949 & 2.67 & 2.53 & \textbf{2.35}  \\ 
 VGG16 & 0.897 & 0.900 & 0.897 & 1.75 & 1.52 & \textbf{1.39}  &0.950 & 0.949 & 0.949 & 2.22 & 2.02 & \textbf{1.87}  \\ 
 Inception & 0.904 & 0.904 & 0.902 & 2.28 & 2.04 & \textbf{1.78}  &0.952 & 0.953 & 0.952 & 3.04 & 2.78 & \textbf{2.55}  \\ 
 ViT & 0.901 & 0.899 & 0.900 & 1.50 & 1.34 & \textbf{1.21}  &0.952 & 0.952 & 0.950 & 1.89 & 1.79 & \textbf{1.58}  \\ 
 CLIP & 0.897 & 0.900 & 0.899 & 1.51 & 1.27 & \textbf{1.16}  &0.949 & 0.949 & 0.950 & 1.89 & 1.57 & \textbf{1.41}  \\ 
\midrule 
 average & 0.899 & 0.900 & 0.898 & 1.97 & 1.79 & \textbf{1.63}  &0.950 & 0.950 & 0.950 & 2.54 & 2.39 & \textbf{2.25}  \\ 
\bottomrule 
\end{tabular}}} 
\end{table}

\section{ESCV on CIAFR-10 and CIFAR-100}
\label{appendix:escv}

\begin{figure}[H]
    \centering
    \resizebox{14cm}{!}{
    \begin{subfigure}{0.5\textwidth}
        \centering
        \includegraphics[width=\linewidth]{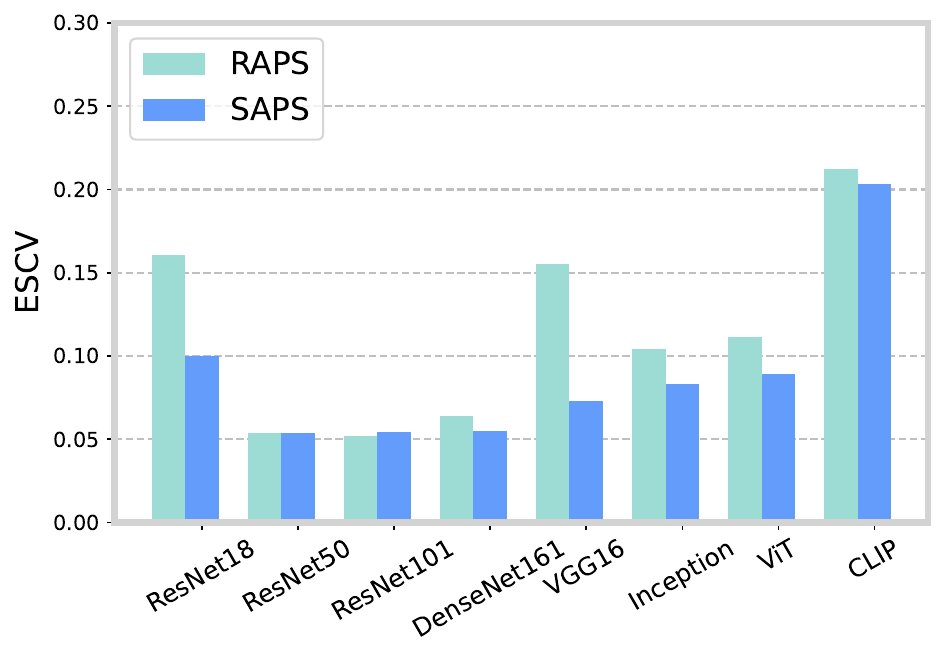}
        \subcaption{CIFAR-10}
        \label{fig:escv_cifar10}
      \end{subfigure}\hfill
      \hspace{1cm}
      \begin{subfigure}{0.5\textwidth}
        \centering
        \includegraphics[width=\linewidth]{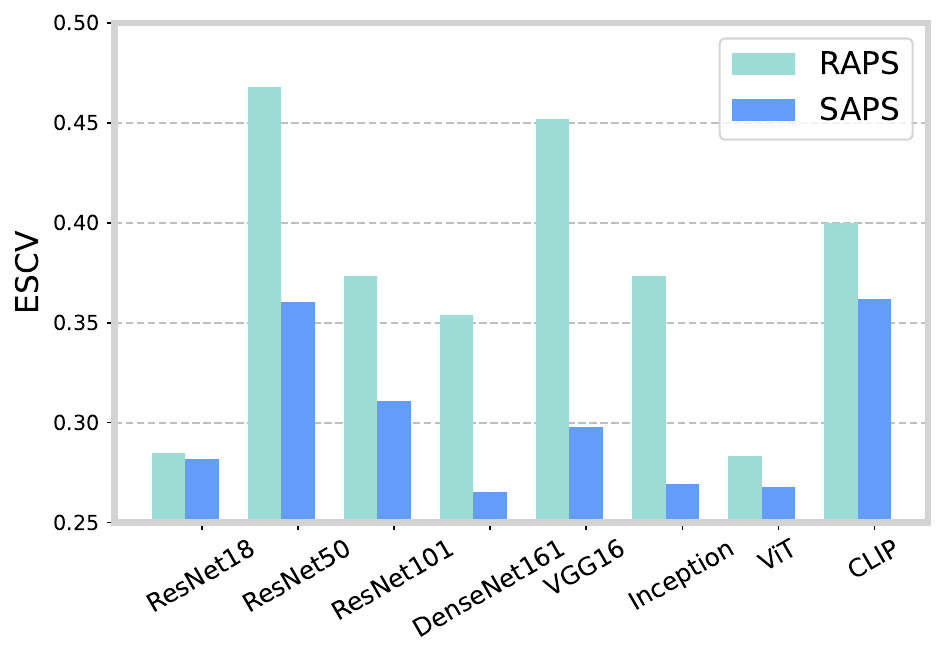}
        \subcaption{CIFAR-100}
        \label{fig:escv_cifar100}
      \end{subfigure}
      }
      \caption{ESCV  for different models on CIFAR-10 and CIFAR-100 with $\alpha=0.1$. }
  \end{figure}

In this section, we report the results of ESCV on CIFAR-10 and CIFAR-100 with $\alpha=0.1$. From Figure~\ref{fig:escv_cifar10} and \ref{fig:escv_cifar100}, we can observe that the SAPS still can get smaller ESCV than RAPS. Moreover, as the scale of the classification task increases, the ESCV for RAPS also exhibits a marked increase, whereas SAPS consistently demonstrates a low value.  For example,  with the CLIP model,  the ESCV for RAPS on ImageNet exceeds that observed for CIFAR-100. But, for both ImageNet and CIFAR-100, the ESCV of SAPS remains around $0.35$.


\section{SAPS vs. RAPS($k_{r}=1$)}
\label{appendix:RAPSK1}

To further understand the influence of excluding the probability value,  we conduct an experiment for RAPS($k_{r}=1$), which is similar to SAPS on ImageNet, CIFAR-100, and CIFAR-10 with $\alpha=0.1$. The median results on ten trials are reported in Table~\ref{table:size_imagenet_raps}, \ref{table:size_cifar100_raps} and \ref{table:size_cifar10_raps}.

\begin{table}[H] 
\centering 
\caption{Comparsion between RAPS($k_{r}=1$) and SAPS  on ImageNet with $\alpha=0.1$. The median-of-means for each column is reported over 10 different trials.  \textbf{Bold} numbers indicate optimal performance.} 
\label{table:size_imagenet_raps} 
\resizebox{0.95\textwidth}{!}{
\setlength{\tabcolsep}{2mm}{
\begin{tabular}{lcccccccccccc} 
\toprule 
 & \multicolumn{6}{c}{$\alpha=0.1$}  & \multicolumn{6}{c}{$\alpha=0.05$} \\ 
\cmidrule(r){2-7}  \cmidrule(r){8-13}  
 & \multicolumn{2}{c}{Coverage}  & \multicolumn{2}{c}{Size $\downarrow$}& \multicolumn{2}{c}{ESCV $\downarrow$} & \multicolumn{2}{c}{Coverage}  & \multicolumn{2}{c}{Size $\downarrow$}  & \multicolumn{2}{c}{ESCV $\downarrow$}  \\ 
\cmidrule(r){2-3} \cmidrule(r){4-5} \cmidrule(r){6-7} \cmidrule(r){8-9} \cmidrule(r){10-11}  \cmidrule(r){12-13}  
Datasets   & \( \begin{array}{@{}c@{}} \text{RAPS} \\ (k_{r}=1) \end{array} \) & SAPS &  \( \begin{array}{@{}c@{}} \text{RAPS} \\ (k_{r}=1) \end{array} \) & SAPS & \( \begin{array}{@{}c@{}} \text{RAPS} \\ (k_{r}=1) \end{array} \) & SAPS & \( \begin{array}{@{}c@{}} \text{RAPS} \\ (k_{r}=1) \end{array} \) & SAPS  & \( \begin{array}{@{}c@{}} \text{RAPS} \\ (k_{r}=1) \end{array} \) & SAPS & \( \begin{array}{@{}c@{}} \text{RAPS} \\ (k_{r}=1) \end{array} \) & SAPS \\ 
\midrule 
 ResNeXt101 & 0.902 & 0.901 & 1.85 & \textbf{1.82}  &0.477 & \textbf{0.366} & 0.950 & 0.950 & 4.56 & \textbf{3.83} & 0.505 & \textbf{0.358}\\ 
 ResNet152 & 0.901 & 0.900 & 1.97 & \textbf{1.92}  &0.572 & \textbf{0.440} & 0.950 & 0.950 & 4.68 & \textbf{4.07} & 0.750 & \textbf{0.335}\\ 
 ResNet101 & 0.900 & 0.900 & 2.09 & \textbf{2.07}  &0.650 & \textbf{0.431} & 0.950 & 0.950 & 4.95 & \textbf{4.34} & 0.724 & \textbf{0.466}\\ 
 ResNet50 & 0.900 & 0.900 & 2.32 & \textbf{2.31}  &0.594 & \textbf{0.350} & 0.950 & 0.950 & 5.52 & \textbf{5.25} & 0.950 & \textbf{0.373}\\ 
 ResNet18 & 0.899 & 0.900 & 4.50 & \textbf{4.00}  &0.900 & \textbf{0.409} & 0.950 & 0.950 & \textbf{10.03} & 10.47 & 0.950 & \textbf{0.490}\\ 
 DenseNet161 & 0.901 & 0.900 & 2.12 & \textbf{2.08}  &0.531 & \textbf{0.393} & 0.951 & 0.951 & 5.15 & \textbf{4.61} & 0.950 & \textbf{0.405}\\ 
 VGG16 & 0.899 & 0.900 & 3.69 & \textbf{3.25}  &0.671 & \textbf{0.425} & 0.949 & 0.949 & \textbf{7.56} & 7.84 & 0.950 & \textbf{0.390}\\ 
 Inception & 0.901 & 0.902 & 5.13 & \textbf{4.58}  &0.478 & \textbf{0.418} & 0.951 & 0.950 & \textbf{13.20} & 14.43 & 0.950 & \textbf{0.532}\\ 
 ShuffleNet & 0.900 & 0.900 & 5.11 & \textbf{4.54}  &0.481 & \textbf{0.375} & 0.951 & 0.950 & \textbf{12.38} & 14.05 & 0.473 & \textbf{0.306}\\ 
 ViT & 0.899 & 0.900 & \textbf{1.58} &1.61  &0.900 & \textbf{0.393} & 0.950 & 0.950 & 3.32 & \textbf{3.21} & 0.950 & \textbf{0.950}\\ 
 DeiT & 0.899 & 0.900 & \textbf{1.38} & 1.41  &0.416 & \textbf{0.389} & 0.950 & 0.949 & 2.59 & \textbf{2.49} & 0.596 & \textbf{0.418}\\ 
 CLIP & 0.900 & 0.900 & 7.16 & \textbf{6.23}  &0.900 & \textbf{0.369} & 0.949 & 0.949 & \textbf{15.16} & 16.07 & 0.950 & \textbf{0.317}\\ 
\midrule 
 average & 0.900 & 0.900 & 3.24 & \textbf{2.98}  &0.631 & \textbf{0.396} & 0.950 & 0.950 & \textbf{7.43} & 7.55&  0.808 & \textbf{0.445}\\ 
\bottomrule 
\end{tabular}}} 
\end{table}

\begin{table}[H] 
\centering 
\caption{Comparsion between RAPS($k_{r}=1$) and SAPS on CIFAR-100 with $\alpha=0.1$. The median-of-means for each column is reported over 10 different trials.  \textbf{Bold} numbers indicate optimal performance.} 
\label{table:size_cifar100_raps} 
\resizebox{0.95\textwidth}{!}{
\setlength{\tabcolsep}{2mm}{
\begin{tabular}{lcccccccccccc} 
\toprule 
 & \multicolumn{6}{c}{$\alpha=0.1$}  & \multicolumn{6}{c}{$\alpha=0.05$} \\ 
\cmidrule(r){2-7}  \cmidrule(r){8-13}  
 & \multicolumn{2}{c}{Coverage}  & \multicolumn{2}{c}{Size $\downarrow$}& \multicolumn{2}{c}{ESCV $\downarrow$} & \multicolumn{2}{c}{Coverage}  & \multicolumn{2}{c}{Size $\downarrow$}  & \multicolumn{2}{c}{ESCV $\downarrow$}  \\ 
\cmidrule(r){2-3} \cmidrule(r){4-5} \cmidrule(r){6-7} \cmidrule(r){8-9} \cmidrule(r){10-11}  \cmidrule(r){12-13}  
Datasets   & \( \begin{array}{@{}c@{}} \text{RAPS} \\ (k_{r}=1) \end{array} \) & SAPS &  \( \begin{array}{@{}c@{}} \text{RAPS} \\ (k_{r}=1) \end{array} \) & SAPS & \( \begin{array}{@{}c@{}} \text{RAPS} \\ (k_{r}=1) \end{array} \) & SAPS & \( \begin{array}{@{}c@{}} \text{RAPS} \\ (k_{r}=1) \end{array} \) & SAPS  & \( \begin{array}{@{}c@{}} \text{RAPS} \\ (k_{r}=1) \end{array} \) & SAPS & \( \begin{array}{@{}c@{}} \text{RAPS} \\ (k_{r}=1) \end{array} \) & SAPS \\ 
\midrule 
 ResNet18 & 0.897 & 0.898 & 2.51 & \textbf{2.41}  &0.321 & \textbf{0.282} & 0.951 & 0.951 & 6.30 & \textbf{4.96} & \textbf{0.209} & 0.244\\ 
 ResNet50 & 0.900 & 0.899 & \textbf{1.97} & 2.04  &0.383 & \textbf{0.360} & 0.948 & 0.946 & 4.52 & \textbf{3.69} & 0.396 & \textbf{0.200}\\ 
 ResNet101 & 0.899 & 0.898 & \textbf{1.92} & 1.99  &0.344 & \textbf{0.311} & 0.950 & 0.948 & 4.56 & \textbf{3.85} & 0.210 & \textbf{0.242}\\ 
 DenseNet161 & 0.899 & 0.897 & \textbf{1.75} & 1.77  &0.381 & \textbf{0.265} & 0.949 & 0.950 & 3.69 & \textbf{3.33} & 0.303 & \textbf{0.298}\\ 
 VGG16 & 0.897 & 0.895 & 4.06 & \textbf{3.64}  &0.567 & \textbf{0.298} & 0.950 & 0.948 & 10.75 & \textbf{9.27} & 0.450 & \textbf{0.325}\\ 
 Inception & 0.901 & 0.902 & \textbf{1.86} & 2.01  &0.375 & \textbf{0.269} & 0.953 & 0.953 & 4.57 & \textbf{4.26} & 0.305 & \textbf{0.213}\\ 
 ViT & 0.902 & 0.900 & \textbf{1.89} & 1.91  &0.312 & \textbf{0.268} & 0.946 & 0.949 & 4.01 & \textbf{3.57} & 0.190 & \textbf{0.170}\\ 
 CLIP & 0.900 & 0.900 & 6.38 & \textbf{5.58}  &0.438 & \textbf{0.362} & 0.949 & 0.950 & \textbf{10.47} & 11.27 & \textbf{0.406} & 0.429 \\ 
\midrule 
 average & 0.899 & 0.899 & 2.79 & \textbf{2.67}  &0.390 & \textbf{0.302} & 0.949 & 0.949 & 6.11 & \textbf{5.53}&  0.309 & \textbf{0.265}\\ 
\bottomrule 
\end{tabular}}} 
\end{table}

\begin{table}[H] 
\centering 
\caption{Comparsion between RAPS($k_{r}=1$) and SAPS on CIFAR-10 with $\alpha=0.1$. The median-of-means for each column is reported over 10 different trials.  \textbf{Bold} numbers indicate optimal performance.}  
\label{table:size_cifar10_raps} 
\resizebox{0.95\textwidth}{!}{
\setlength{\tabcolsep}{2mm}{
\begin{tabular}{lcccccccccccc} 
\toprule 
 & \multicolumn{6}{c}{$\alpha=0.1$}  & \multicolumn{6}{c}{$\alpha=0.05$} \\ 
\cmidrule(r){2-7}  \cmidrule(r){8-13}  
 & \multicolumn{2}{c}{Coverage}  & \multicolumn{2}{c}{Size $\downarrow$}& \multicolumn{2}{c}{ESCV $\downarrow$} & \multicolumn{2}{c}{Coverage}  & \multicolumn{2}{c}{Size $\downarrow$}  & \multicolumn{2}{c}{ESCV $\downarrow$}  \\ 
\cmidrule(r){2-3} \cmidrule(r){4-5} \cmidrule(r){6-7} \cmidrule(r){8-9} \cmidrule(r){10-11}  \cmidrule(r){12-13}  
Datasets   & \( \begin{array}{@{}c@{}} \text{RAPS} \\ (k_{r}=1) \end{array} \) & SAPS &  \( \begin{array}{@{}c@{}} \text{RAPS} \\ (k_{r}=1) \end{array} \) & SAPS & \( \begin{array}{@{}c@{}} \text{RAPS} \\ (k_{r}=1) \end{array} \) & SAPS & \( \begin{array}{@{}c@{}} \text{RAPS} \\ (k_{r}=1) \end{array} \) & SAPS  & \( \begin{array}{@{}c@{}} \text{RAPS} \\ (k_{r}=1) \end{array} \) & SAPS & \( \begin{array}{@{}c@{}} \text{RAPS} \\ (k_{r}=1) \end{array} \) & SAPS \\ 
\midrule 
 ResNet18 & 0.897 & 0.898 & 2.24 & \textbf{2.18}  &0.145 & \textbf{0.100} & 0.946 & 0.949 & \textbf{2.98} & 3.10 & \textbf{0.052} & 0.076 \\ 
 ResNet50 & 0.899 & 0.895 & 1.79 & \textbf{1.77}  &0.101 & \textbf{0.054} & 0.949 & 0.950 & 2.63 & \textbf{2.53} & 0.062 & \textbf{0.030}\\ 
 ResNet101 & 0.899 & 0.898 & 1.87 & \textbf{1.86}  &0.107 & \textbf{0.054} & 0.946 & 0.949 & 2.61 & \textbf{2.59} & 0.043 & \textbf{0.026}\\ 
 DenseNet161 & 0.899 & 0.898 & \textbf{1.69} & 1.71  &0.156 & \textbf{0.055} & 0.952 & 0.949 & 2.52 & \textbf{2.35} & 0.103 & \textbf{0.021}\\ 
 VGG16 & 0.900 & 0.897 & \textbf{1.38} & 1.39  &0.112 & \textbf{0.073} & 0.949 & 0.949 & \textbf{1.84} & 1.87 & 0.124 & \textbf{0.050}\\ 
 Inception & 0.903 & 0.902 & \textbf{1.77} & 1.78  &0.143 & \textbf{0.083} & 0.954 & 0.952 & 2.68 & \textbf{2.55} & 0.089 & \textbf{0.040}\\ 
 ViT & 0.900 & 0.900 &\textbf{ 1.16} & 1.21  &0.127 & \textbf{0.089} & 0.950 & 0.950 & 1.63 & \textbf{1.58} & 0.101 & \textbf{0.054}\\ 
 CLIP & 0.899 & 0.899 & \textbf{1.06} & 1.16  &0.213 & \textbf{0.203} & 0.950 & 0.950 & \textbf{1.36} & 1.41 & 0.254 & \textbf{0.123}\\ 
\midrule 
 average & 0.900 & 0.898 & \textbf{1.62} & 1.63  &0.138 & \textbf{0.089} & 0.949 & 0.950 & 2.28 & \textbf{2.25}&  0.104 & \textbf{0.052}\\ 
\bottomrule 
\end{tabular}}} 
\end{table}

\end{document}